\documentclass[10pt,twocolumn,letterpaper]{article}

\usepackage[pagenumbers]{cvpr} 

\usepackage{amsmath,amsfonts,bm}

\def\eqref#1{equation~\ref{#1}}

\def\1{\bm{1}}

\def\vx{{\bm{x}}}
\def\vy{{\bm{y}}}
\def\vz{{\bm{z}}}

\DeclareMathAlphabet{\mathsfit}{\encodingdefault}{\sfdefault}{m}{sl}
\SetMathAlphabet{\mathsfit}{bold}{\encodingdefault}{\sfdefault}{bx}{n}

\newcommand{\E}{\mathbb{E}}

\usepackage[accsupp]{axessibility}  

\usepackage{graphicx}
\usepackage{amsmath}
\usepackage{amssymb}
\usepackage{booktabs}

\usepackage{multirow}
\usepackage{xcolor} 
\usepackage{colortbl} 
\usepackage{balance} 

\usepackage[pagebackref,breaklinks,colorlinks]{hyperref}

\usepackage[capitalize]{cleveref}
\crefname{section}{Sec.}{Secs.}
\Crefname{section}{Section}{Sections}
\Crefname{table}{Table}{Tables}
\crefname{table}{Tab.}{Tabs.}

\def\cvprPaperID{2991} 
\def\confName{CVPR}
\def\confYear{2022}

\begin{document}

\title{ELIC: Efficient Learned Image Compression with \\ Unevenly Grouped Space-Channel Contextual Adaptive Coding}

\newcommand{\authorinstitude}[1]{{\textsuperscript{#1}}}

\author{Dailan He\authorinstitude{1}\thanks{Equal contribution.}\ ,
Ziming Yang\authorinstitude{1}\footnotemark[1]\ , 
Weikun Peng\authorinstitude{1}, 
Rui Ma\authorinstitude{1},
Hongwei Qin\authorinstitude{1}, 
Yan Wang\authorinstitude{1}\authorinstitude{2}\thanks{Corresponding author. This work is done when Ziming Yang, Weikun Peng and Rui Ma are interns at SenseTime Research.}\\
SenseTime Research\authorinstitude{1}, Tsinghua University\authorinstitude{2}\\
{\tt\small \{hedailan, yangziming, pengweikun, marui, qinhongwei, wangyan1\}@sensetime.com} \\
{\tt\small wangyan@air.tsinghua.edu.cn}
}
\maketitle

\begin{abstract}
Recently, learned image compression techniques have achieved remarkable performance, even surpassing the best manually designed lossy image coders. They are promising to be large-scale adopted. For the sake of practicality, a thorough investigation of the architecture design of learned image compression, regarding both compression performance and running speed, is essential.
In this paper, we first propose uneven channel-conditional adaptive coding, motivated by the observation of energy compaction in learned image compression. Combining the proposed uneven grouping model with existing context models, we obtain a spatial-channel contextual adaptive model to improve the coding performance without damage to running speed. Then we study the structure of the main transform and propose an efficient model, ELIC, to achieve state-of-the-art speed and compression ability. With superior performance, the proposed model also supports extremely fast preview decoding and progressive decoding, which makes the coming application of learning-based image compression more promising.
\end{abstract}

\section{Introduction}
\label{sec:intro}

In the past years, lossy image compression based on deep learning develops rapidly~\cite{balle2016end, balle2018variational, minnen2018joint, lee2019hybrid, cheng2020learned, minnen2020channel, he2021checkerboard,guo2021causal, xie2021inn, gao2021attn-back-proj-iccv, wu2021block}. They have achieved remarkable progress on improving the rate-distortion performance, with usually much better MS-SSIM~\cite{wang2003multiscale} than conventional image formats like JPEG~\cite{JPEG-ITU1992Information} and BPG~\cite{bellard2015bpg}, which indicates better subjective quality. Some very recent works~\cite{guo2021causal, xie2021inn, gao2021attn-back-proj-iccv, wu2021block, egilmez2021transform, fu2021GLLMM} even outperform the still image coding of VVC~\cite{vtm2019}, one of the best hand-crafted image and video coding standards at present, on both PSNR and MS-SSIM. These results are encouraging, as learned image compression has been proved as a strong candidate for next generation image compression techniques. In the near future, it is quite possible to deploy this line of image compression models in industrial applications. Yet, to make these approaches practical, we must carefully assess the running speed, especially the decoding speed of learned image compression.

\begin{figure}[t]
    \centering
    \subfloat{
    \includegraphics[width=0.90\linewidth]{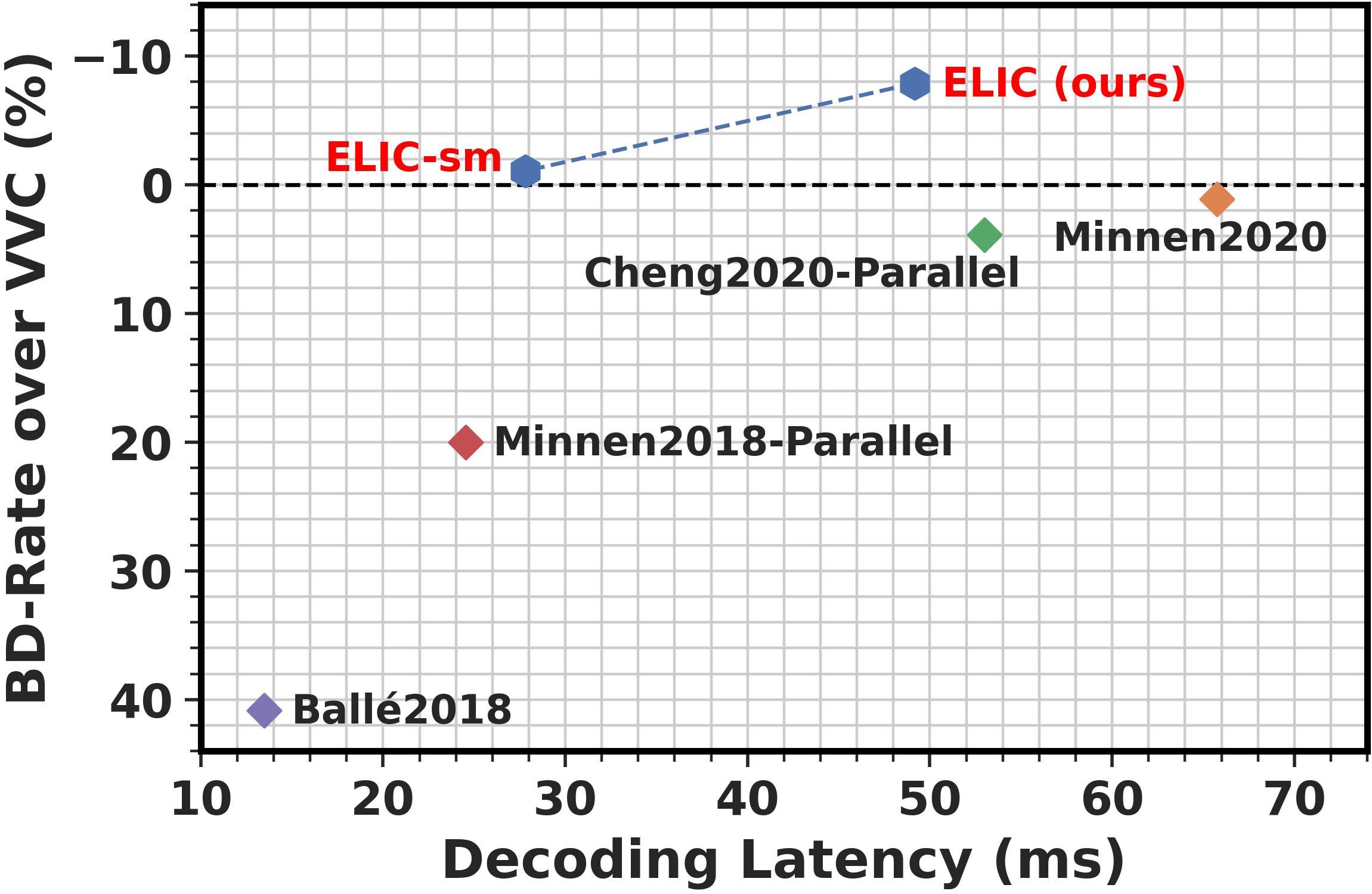}
    }
    \caption{Rate-speed comparison on Kodak. Left-top is better. 
    }
    \label{fig:comp-bdbr-speed}
\end{figure}

One of the most important techniques in learned image compression is the joint backward-and-forward adaptive entropy modeling~\cite{weinlich2016probability, minnen2020channel,minnen2018joint, lee2019hybrid, cheng2020learned, guo2021causal, xie2021inn, gao2021attn-back-proj-iccv, wu2021block}. It helps convert the marginal probability model of coding-symbols to a joint model by introducing extra latent variables as prior~\cite{balle2018variational, minnen2018joint, minnen2020channel}, leading to less redundancy and lower bit-rate.  However, the backward-adaptive models along spatial dimension significantly break the parallelism, which inevitably slows down the decoding. To address the issue, He~\etal~\cite{he2021checkerboard} proposes to adopt checkerboard convolution as a parallel replacement to the serial autoregressive context model, which has a much better degree of parallelism with constant complexity. Minnen~\etal~\cite{minnen2020channel} proposes to adopt a context model along channel dimension instead of the serial-decoded spatial ones, which also improves the parallelism. However, to achieve a non-trivial bit-saving with this channel-conditional model, the symbols are divided to 10 groups and coded progressively,  which still slows down the overall inference. It is promising to delve into parallel multi-dimension contextual adaptive coding by combing these two models to achieve better coding ability~\cite{he2021checkerboard}, constituting one of the motivation of our work. In this paper, we investigate an uneven grouping scheme to speed up the channel-conditional method, and further combine it with a parallel spatial context model, to promote RD performance while keeping a fast running speed.

More and more complex transform networks also slow down the inference. As learned image compression is formulated as a sort of nonlinear transform coding~\cite{goyal2001theoretical-foundations, balle2020nonlinear},  another plot improving coding performance is the development of main transform. Prior works introduce larger networks~\cite{cheng2020learned, lin2020blocklstm, guo2021causal, gao2021attn-back-proj-iccv}, attention modules~\cite{liu2020edic, liu2019non, cheng2020learned, guo2021causal} or invertible structures~\cite{ma2020iwave++, xie2021inn} to main analysis and synthesis networks. These heavy structures significantly improve the RD performance but hurt the speed. We notice that, with a relative strong and fast adaptive entropy estimation (\ie the above mentioned adaptive coding approaches with hyperprior and context model), we can re-balance the computation between main transform and entropy estimation, to obtain low-latency compression models. This further motivates us to promote the contextual modeling technique. 

Learned image compression is growing mature and tends to be widely used, but its lack of efficiency is still a critical issue. In this paper, we contribute to this field from following perspectives:
\begin{itemize}
    \item We introduce information compaction property as an inductive bias to promote the expensive channel-conditional backward-adaptive entropy model. Combining it with the spatial context model,  we propose a multi-dimension entropy estimation model named Space-Channel ConTeXt (SCCTX) model, which is fast and effective on reducing the bit-rate.
    \item  With the proposed SCCTX model, we further propose ELIC model. The model adopts stacked residual blocks as nonlinear transform, instead of GDN layers~\cite{balle2016end}. It surpasses VVC on both PSNR and MS-SSIM, achieving state-of-the-art performance regarding coding performance and running speed (Figure~\ref{fig:comp-bdbr-speed} and Table~\ref{tab:comp-full-table}).
    \item We propose an efficient approach to generate preview images from the compressed representation. To our knowledge, this is the first literature addressing the very-fast preview issue of learned image compression.
\end{itemize}

\section{Related works}

\subsection{Learned lossy image compression}

Learned lossy image compression~\cite{balle2018variational, minnen2018joint, liu2020edic, cheng2020learned, he2021checkerboard, guo2021causal} aims at establishing a data-driven rate-distortion optimization (RDO) approach. Given input image $\vx$ and a pair of neural analyzer $g_a$ and neural synthesizer $g_s$, this learning-based RDO is formulated as: 
\begin{equation}
\begin{split}
    \mathcal{L} &= R(\hat \vy) + \lambda D(\vx, g_s(\hat \vy))
\end{split}
\label{eq:learned-rdo}
\end{equation}
where $\hat \vy = \lceil g_a(\vx) \rfloor$ represents the discrete coding-symbols to be saved and $\lceil\cdot\rfloor$ is the quantization operator. Balancing the estimated bit-rate $R$ and image reconstruction distortion $D$ with a rate-controlling hyper-parameter $\lambda$, we can train a set of neural networks $g_a, g_s$ to obtain various pairs of image en/de-coding models, producing a rate-distortion curve.

Ball\'e \etal~\cite{balle2016end} proposes to adopt a uniform noise estimator and a parametric entropy model to approximate the probability mass function $p_{\hat\vy}$, so that its expected negative entropy $-\E [\log p_{\hat\vy}(\hat \vy)]$ can be  supervised as the $R(\hat \vy)$ term in eq.~\ref{eq:learned-rdo} in a differentiable manner with gradient-decent-based optimization. Later, the entropy model is further extended to a conditioned Gaussian form~\cite{balle2018variational, minnen2018joint}:
\begin{equation}
    p_{\hat\vy|\hat\vz}(\hat\vy | \hat \vz) = \left[\mathcal{N}(\bm{\mu}, \bm{\sigma}^2) * U(-0.5, 0.5)\right] (\hat \vy)
\end{equation}
where the entropy parameters $\bm{\Theta} = (\bm{\mu}, \bm{\sigma}^2)$ are calculated from extra computed or stored prior. Ball\'e \etal~\cite{balle2018variational} adopts hyperprior $\hat \vz$ to calculate the entropy parameters. $\hat \vz$ is calculated from unquantized symbols $\vy$ with a hyper analyzer $h_a$. It can be seen as side-information introduced to the neural coder, acting as the forward-adaptive method.

To painlessly improve the coding efficiency, several training, inference, and encoding-time optimizing approaches are proposed~\cite{yang2020improving-inference, zhao2021encoding-opt, guo2021soft-then-hard}. They can improve the RD performance without slow down the decoding, and can be used together with various coding architectures. 

\newcommand{\energyvisimgwidth}{0.08\linewidth}
\begin{figure*}[]
    \centering
    \subfloat{%
       \includegraphics[width=\energyvisimgwidth]{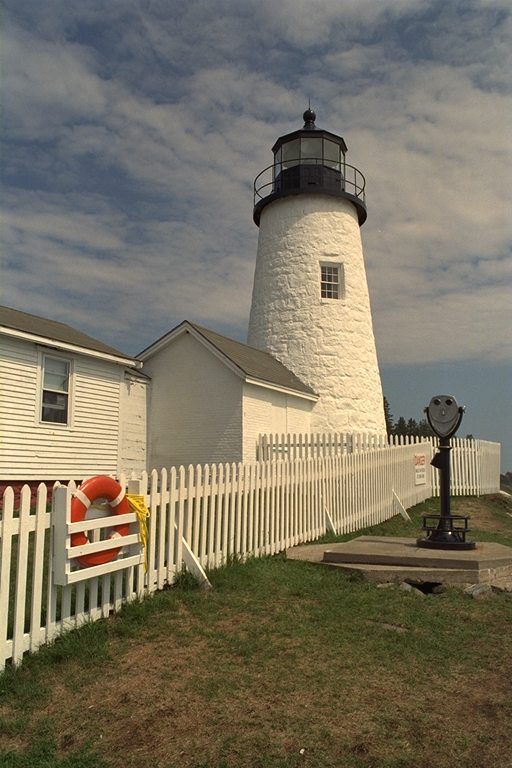}}
    \hfill
    \subfloat{%
       \includegraphics[width=\energyvisimgwidth]{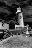}}
    \hfill
  \subfloat{%
        \includegraphics[width=\energyvisimgwidth]{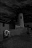}}
    \hfill
  \subfloat{%
        \includegraphics[width=\energyvisimgwidth]{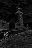}}
    \hfill
  \subfloat{%
        \includegraphics[width=\energyvisimgwidth]{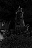}}
    \hfill
\subfloat{%
        \includegraphics[width=\energyvisimgwidth]{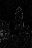}}
    \hfill
\subfloat{%
        \includegraphics[width=\energyvisimgwidth]{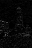}}
    \hfill
\subfloat{%
        \includegraphics[width=\energyvisimgwidth]{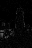}}
    \hfill
\subfloat{%
        \includegraphics[width=\energyvisimgwidth]{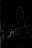}}
    \hfill
\subfloat{%
        \includegraphics[width=\energyvisimgwidth]{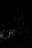}}
    \hfill
\subfloat{%
        \includegraphics[width=\energyvisimgwidth]{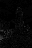}}
    \caption{Visualization of sorted channels with the top-10 largest average energy. Left 1: the original image (\texttt{kodim19.png}). Lighter regions correspond to larger symbol magnitudes $|\hat y|$. It can been seen from the figures that most strong activation concentrates in the first channel (left 2) and the remained channels become sparse gradually.}
    \label{fig:vis-energy-compact}
\end{figure*}

\subsection{Backward-adaptive entropy models}

Backward-adaptive coding is also introduced to learned image compression, including spatial context models~\cite{minnen2018joint, lee2019hybrid, he2021checkerboard, guo2021causal} and channel-conditional models~\cite{minnen2020channel}. Correlating current decoding symbols with already decoded symbols, this sort of approaches further save the bits.

A spatial context model refers to observable neighbors of each symbol  vector $\hat \vy_i$ at the $i$-th location:
\begin{align}
    \hat \vy_{<i} &= \{\hat \vy_1, \dots, \hat \vy_{i-1}\} \\
    \Phi_{\mathrm{sp},i} &= g_{\mathrm{sp}}(\hat \vy_{<i}) \label{eq:sp-ctx}
\end{align}
where the context model $g_{\mathrm{sp}}(\cdot)$ is an autoregressive convolution~\cite{oord2016pixelcnn, minnen2018joint}. Each context representation $\Phi_{\mathrm{sp},i}$ is used to jointly predict entropy parameters accompanied by the hyperprior $\hat \vz$. This approach demands symbol vectors $\hat \vy_1, \dots, \hat \vy_{HW}$ to be decoded serially, which critically slows down the decoding~\cite{minnen2018joint, liu2020edic, he2021checkerboard}.  He \etal~\cite{he2021checkerboard} proposes to separate the symbols into anchors and non-anchors:
\begin{align}
    \hat \vy^{(\mathrm{anchor})}_{<i} = \varnothing, \quad \hat \vy^{(\mathrm{nonanc})}_{<i} = \hat \vy^{(\mathrm{anchor})}
\end{align}
and adopts a checkerboard convolution as $g_{\mathrm{sp}}(\cdot)$, so  the decoding of both anchors and non-anchors can be in parallel.

Another scheme to perform parallel backward adaption is to reduce the redundancy among channels. Minnen \etal~\cite{minnen2020channel} proposes to group the symbol channels to $K$ chunks as the channel-wise context:
\begin{align}
    \Phi_{\mathrm{ch}}^{(k)} &= g_{\mathrm{ch}}^{(k)}(\hat \vy^{<k}),\quad k=2,\dots, K
\label{eq:cc}
\end{align}
where $\hat \vy^{<k} = \{\hat \vy^{(1)},\dots, \hat \vy^{(k-1)}\}$ denotes already decoded channel chunks. Setting a proper $K$ is essential for this method to balance the compression performance and running speed. The larger $K$ is, the better the RD performance is~\cite{minnen2020channel}, yet the slower the parameter estimation is (as the degree of parallelism decreases).

Multi-dimension adaptive coding approaches have been proposed but all of them still suffer from the slow-speed issue, to our knowledge. Liu~\etal~\cite{liu2019non} proposes a 3D context model which performs a 3D convolution passing by all the channels. Li~\etal~\cite{li2020efficient} proposes a multi-dimension context model with non-constant complexity. Guo~\etal~\cite{guo2021causal} uses 2-chunk contextual modeling with serial global-and-local adaptive coding. Ma~\etal~\cite{ma2021cross} propose a cross-channel context model which uses a even denser referring scheme than 3D context models.

\section{Parallel multi-dimension context modeling}

\begin{figure}[t]
    \centering
    \includegraphics[width=\linewidth]{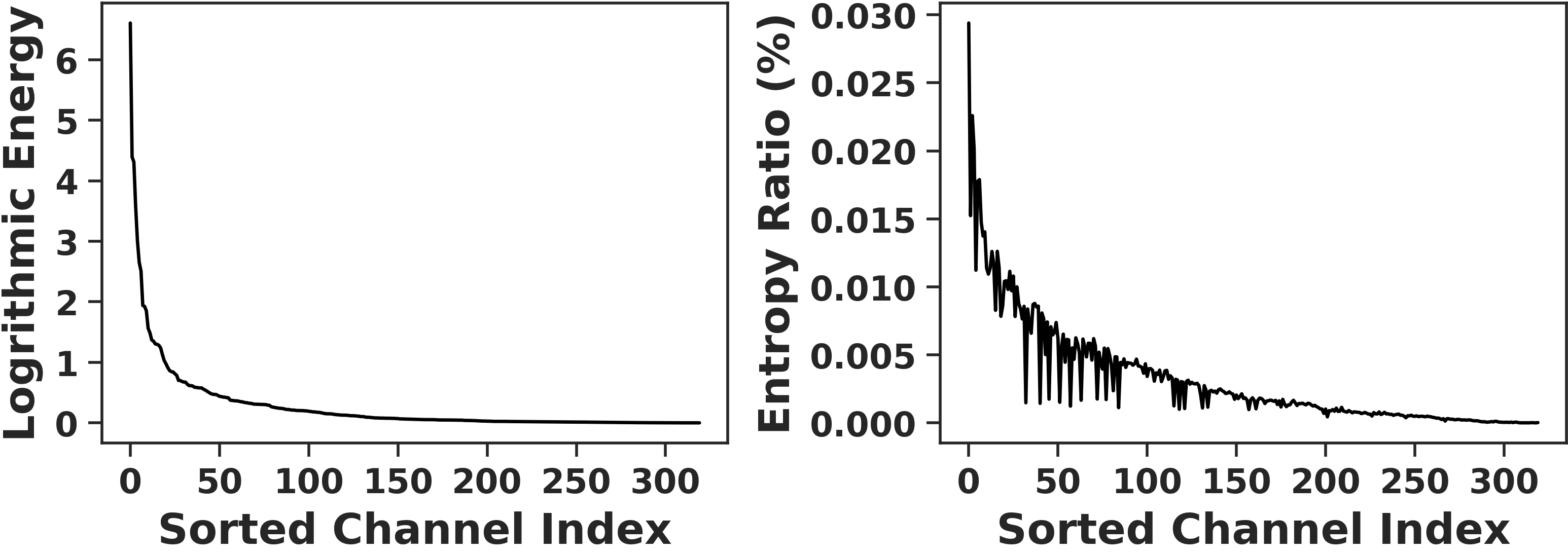}
    \caption{Energy and entropy distribution along channels. The results are evaluated with Minnen2018~\cite{minnen2018joint} model, on Kodak. The channels are sorted by energy averaged over all 24 images.}
    \label{fig:dist-energy-entropy}
\end{figure}

\subsection{Information compaction property}
Energy compaction is an essential property of transform coding~\cite{goyal2001theoretical-foundations}, \eg DCT-based JPEG. With decomposed coefficients extremely concentrated on lower frequencies, describing most structural and semantic information of the original image, higher frequencies can be compressed more heavily by using larger quantization steps to achieve a better rate-distortion trade-off.

We find this compaction also exists in learned analysis transform. We visualize each latent feature map $\hat \vy^{(\ell)}$ of the mean-scale joint model, Minnen2018~\cite{minnen2018joint}, as its scaled magnitude (Figure~\ref{fig:vis-energy-compact})
and draw the distribution of energy and entropy along channels (Figure~\ref{fig:dist-energy-entropy}). More strongly activated, the beginning channels have much larger average energy. Since the entropy distribution correlates to the energy distribution, it indicates an information compaction property. 
This phenomenon exists in all of the 5 models we test: Ball\'e2018~\cite{balle2018variational}, Minnen2018~\cite{minnen2018joint}, Cheng2020~\cite{cheng2020learned} and their parallel versions~\cite{he2021checkerboard}. The information compaction in those models is orderless because the supervision conducted on the analyzer output channels is symmetric. 

\newcommand{\infoorderheight}{0.35\linewidth}
\begin{figure}[t]
    \centering
    \subfloat[Entropy of each channel.]{
    \includegraphics[height=\infoorderheight]{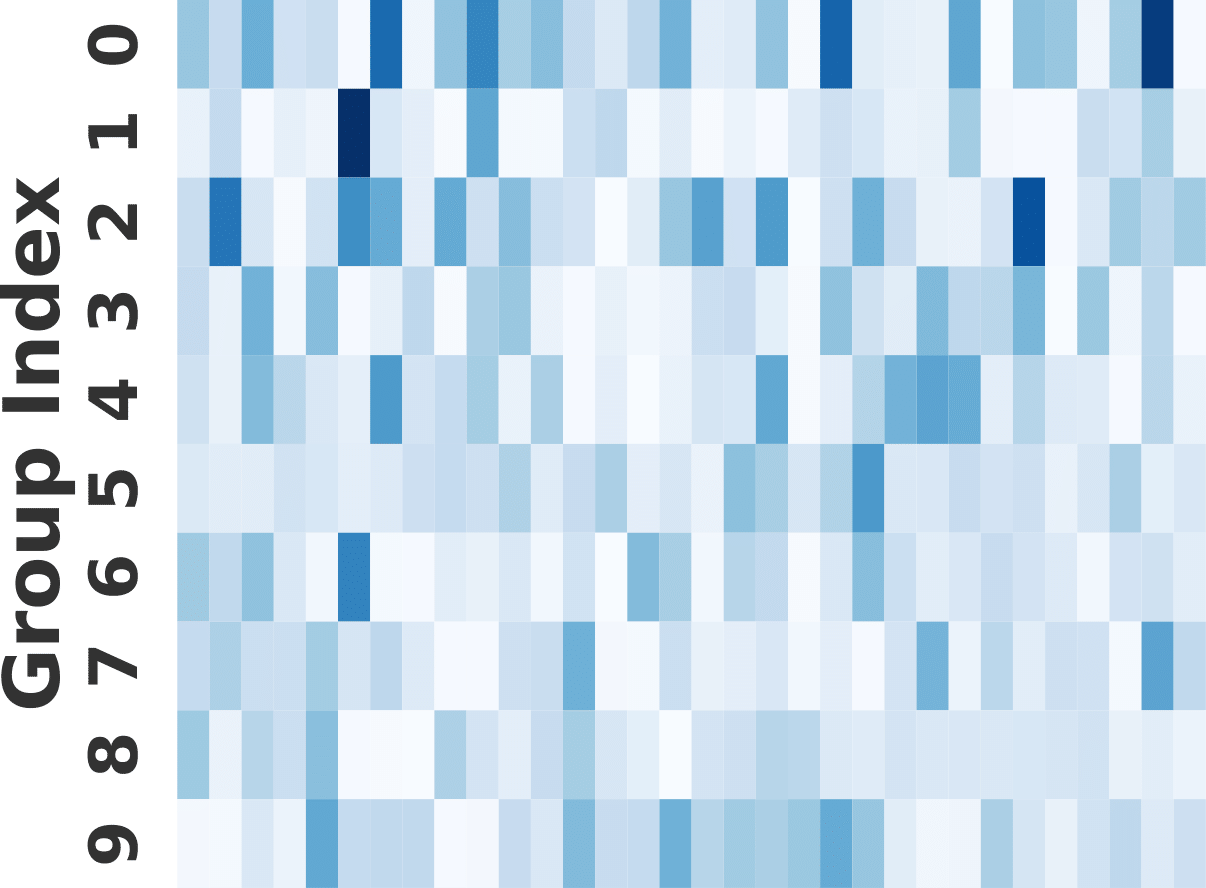}
    }
        \subfloat[Maximal entropy of each group.]{
    \includegraphics[height=\infoorderheight]{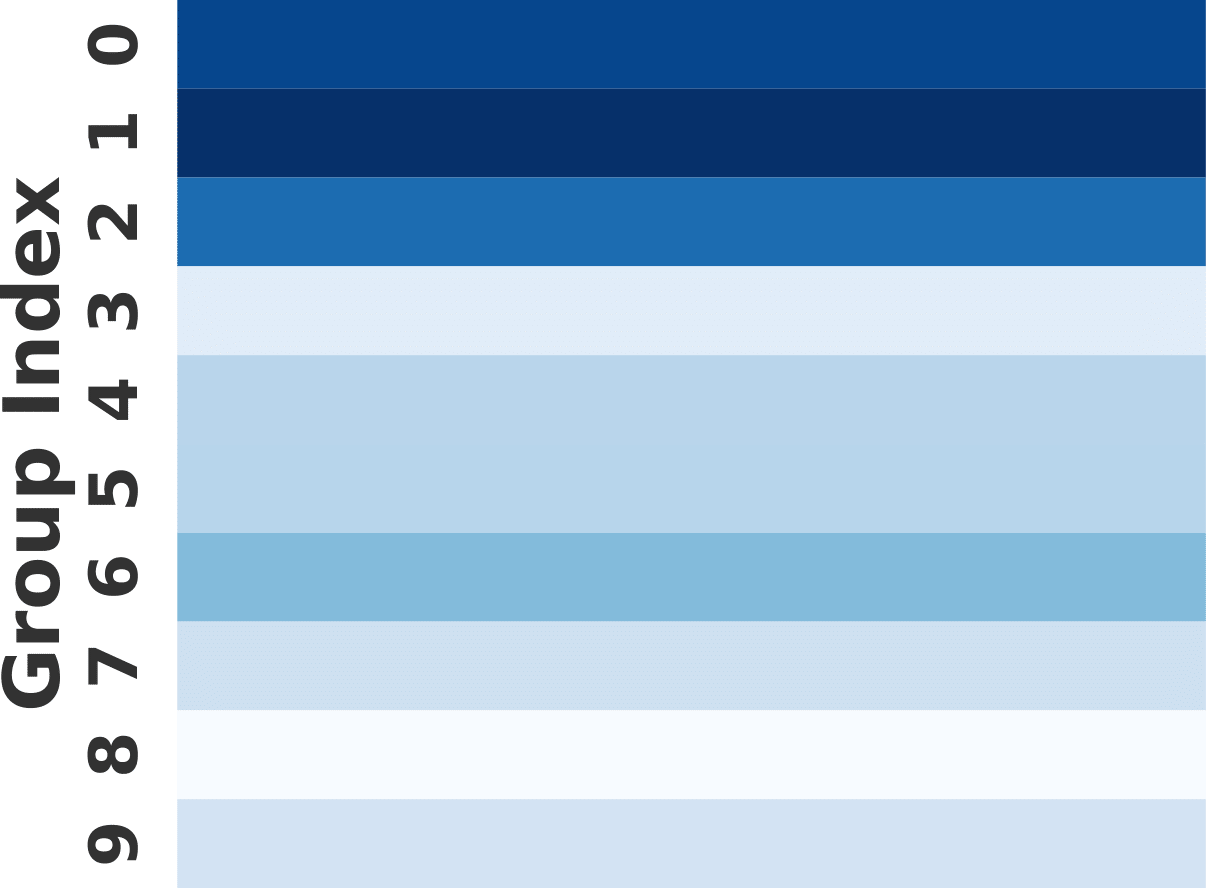}
    }
    \caption{A case study of adopting the 10-slice channel-conditional adaptive coding~\cite{minnen2020channel}. Deeper colors denote larger values. The entropy of each channel group is implicitly sorted. The beginning groups contain channels with the largest entropy.
    The results are from evenly grouped model trained with $\lambda=0.045$, evaluated on \texttt{kodim08} from Kodak. }
    \label{fig:vis-channel-sort}
\end{figure}

When adopting a channel-conditional approach, this property induces a group-level order. See Figure~\ref{fig:vis-channel-sort}. Particular channels in the earlier encoded groups have much larger entropy, so they are allocated more bits. As the beginning channels are more frequently referred to by following channels, the major information implicitly concentrates on the beginning channels to help eliminate more channel-wise redundancy. The progressive coding results~\cite{minnen2020channel} also experimentally prove this, since we can reconstruct the main semantics of images only from the beginning channels.

We tend to understand this information compaction property of learned image compression from the perspective of sparse representation learning~\cite{ranzato2007efficient-learning-of-sparse-representations, ranzato2007sparse, ng2011sparse-autoencoder}, yet the theoretical analysis and explanation of it are beyond the topic of this paper. We view it as a strong prior knowledge that helps us introduce inductive bias to improve the model design.

\subsection{Unevenly grouped channel-wise context model}
\label{sec:unevenly-grouping}

\begin{figure}[t]
    \centering
    \includegraphics[width=0.98\linewidth]{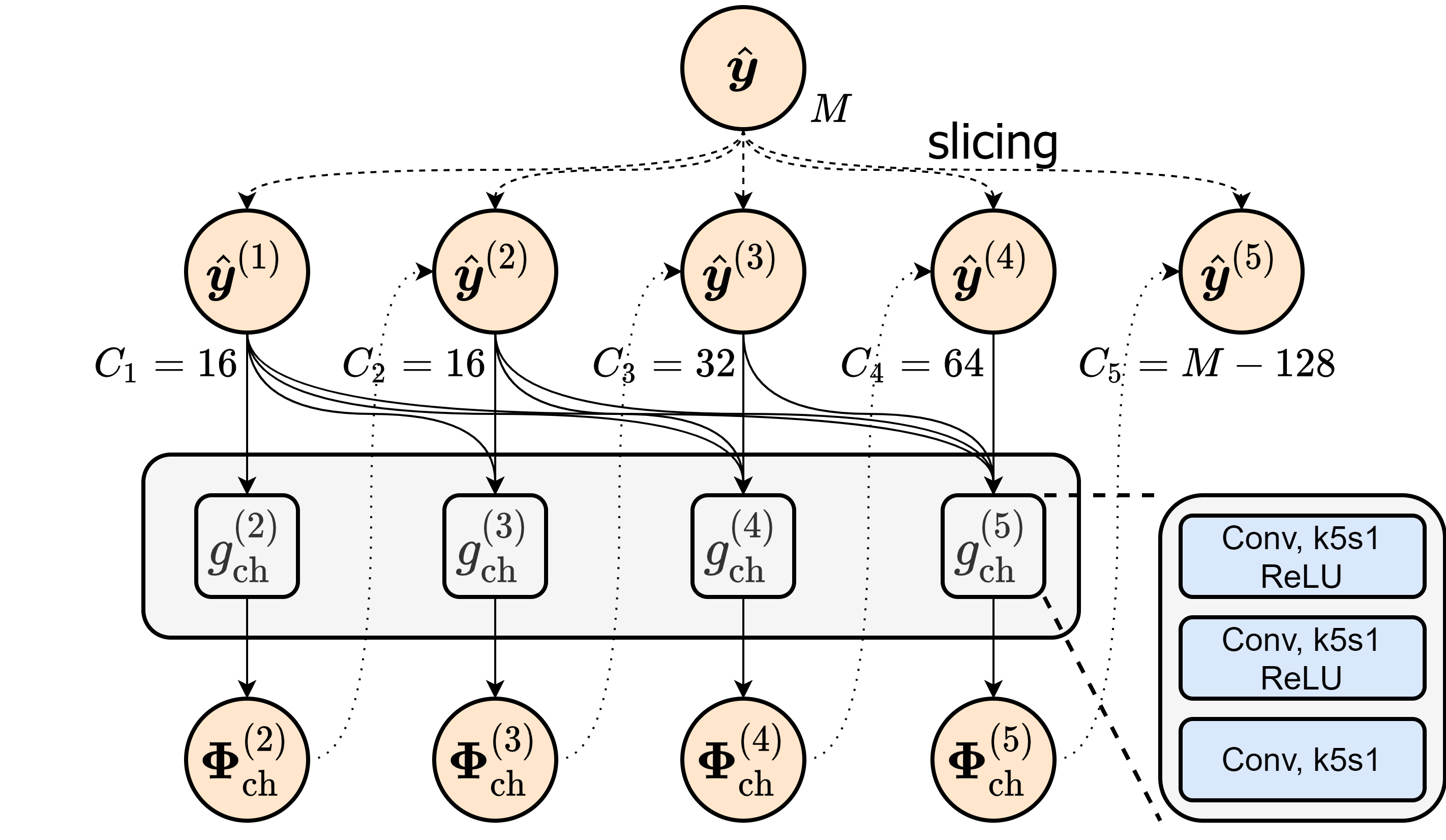}
    \caption{Proposed uneven grouping for channel-conditional (CC) adaptive coding. The $M$-channel coding-symbols $\hat \vy$ are grouped into 5 chunks with gradually increased number of channels $C_k$.}
    \label{fig:unevenly-cc}
\end{figure}

The visualization in Figure~\ref{fig:vis-channel-sort} shows that the later encoded channels contain less information, and they are less frequently used to predict following groups. Therefore, we can reduce the cross-group reference to speed up, by merging more later encoded channels into larger chunks. On the other hand, because of the information compaction, with less channel, the earlier encoded channel groups may still well help reduce the entropy of the following channels. Thus, a more elaborate channel grouping scheme may further improve this entropy estimation module by re-balancing the channel numbers of different groups. Yet, existing approaches often simply group the channels to chunks with the same size~\cite{minnen2020channel, guo2021causal} or adopt per-channel grouping~\cite{liu2019non, ma2021cross}.

We propose an uneven grouping scheme, allocating finer granularity to the beginning chunks by using fewer channels and grow coarser gradually for the following chunks by using more channels. Thus, for symbols $\hat \vy$ with $M$ channels, we split them along the channel dimension to 5 chunks $ \hat\vy^{(1)}, \dots, \hat\vy^{(5)}$ with $16, 16, 32, 64, M-128$ channels respectively. Figure~\ref{fig:unevenly-cc} shows the channel-conditional model with this uneven allocation. It only requires $5$ times of parallel calculation to decode all the slices $\hat \vy^{(k)}$, which saves the running time. 

\begin{figure}[t]
    \centering
    \includegraphics[width=0.95\linewidth]{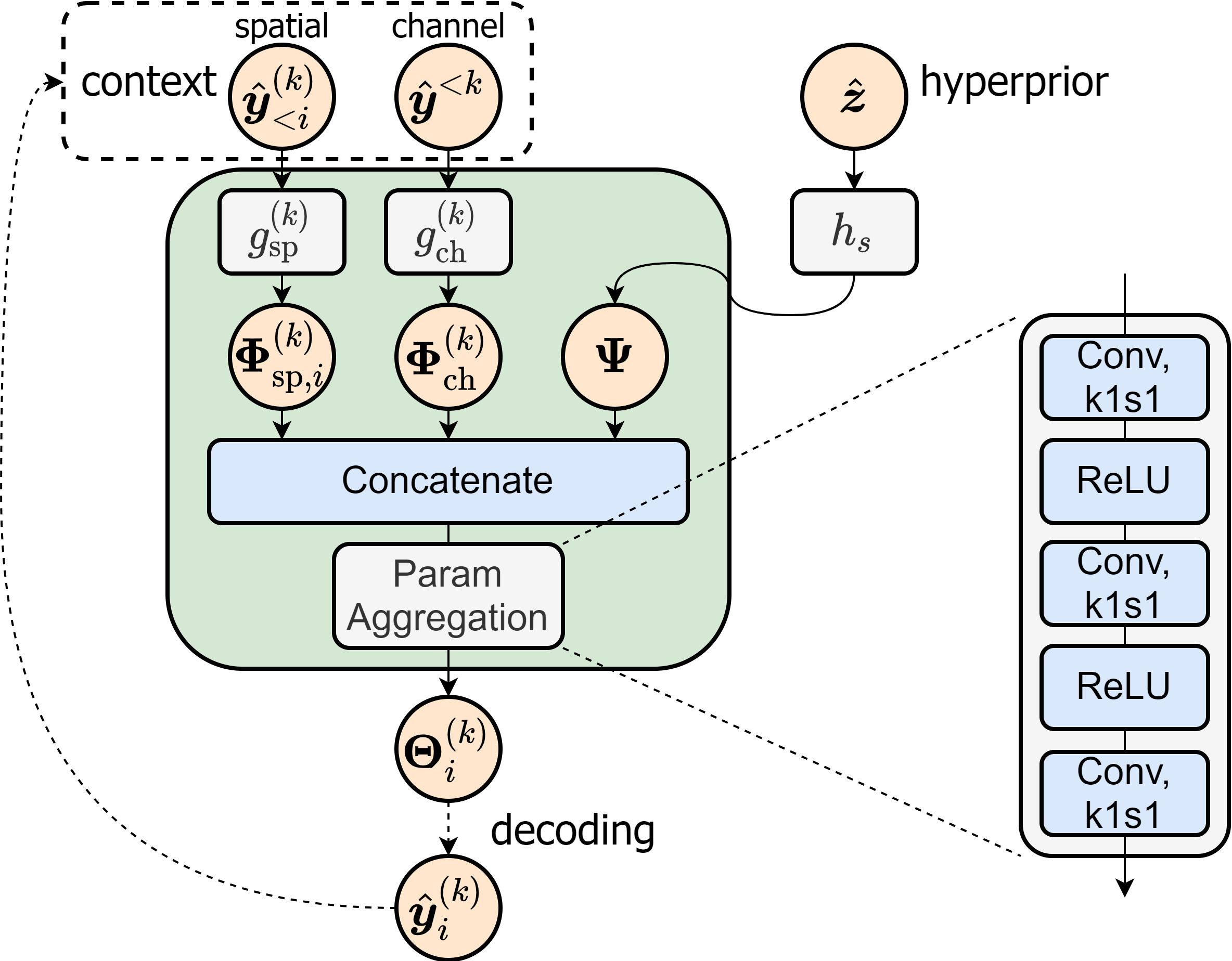}
    \caption{Diagram of proposed space-channel context model.}
    \label{fig:scctx}
\end{figure}

\subsection{SCCTX: space-channel context model}
\label{sec:scctx}

\begin{table}[ht]
    \centering
    \begin{tabular}{lcrr}
    \toprule
         \textbf{Method} & \textbf{Type} & $G$ & $S$ ($M=320$) \\
    \midrule
         autoregressive~\cite{minnen2018joint} & sp. & $HW$ & $320$ \\
         checkerboard~\cite{he2021checkerboard} & sp. & 2 & $160 HW$\\
         10-slice CC.~\cite{minnen2020channel}& ch. & 10 & $32 HW$ \\
         uneven CC. (ours) & ch. & 5 & $(16\to 192)HW$ \\
         SCCTX (ours)& mix & 10 & $(8\to 96)HW$ \\
    \bottomrule
    \end{tabular}
    \caption{Comparison of different backward-adaptive coding approaches. $G$ denotes the number of groups, and $S$ is the group size. Symbols in the same group can be processed in parallel. The sizes are calculated for $M\times H\times W$ symbols where $M=320$. 
    }
    \label{tab:backward-adaption-schemes}
\end{table}

\begin{figure*}[htb]
    \centering
    \includegraphics[width=0.98\linewidth]{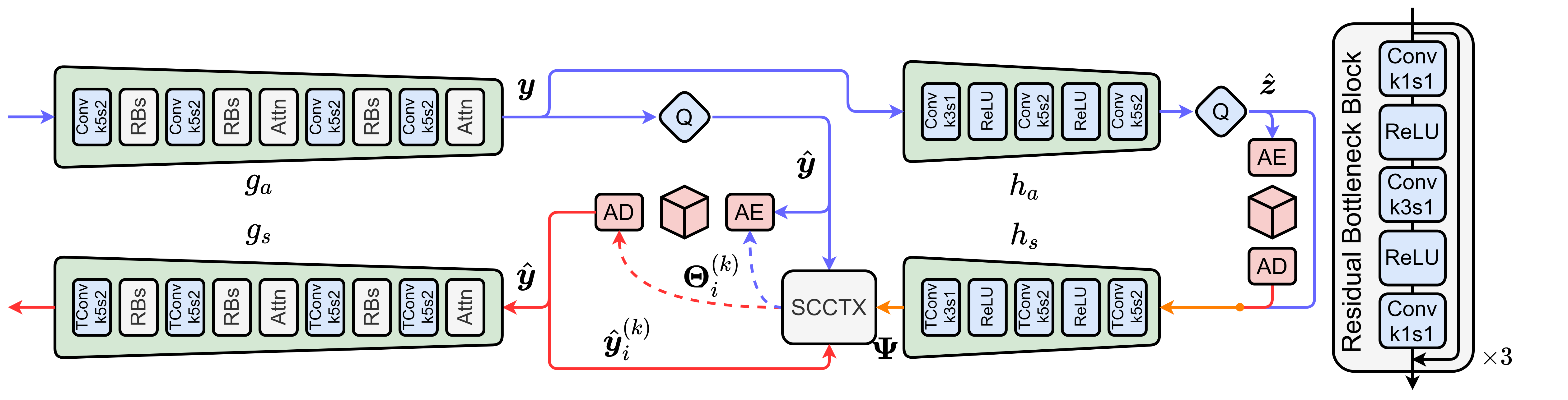}
    \caption{The overall architecture of ELIC. \textit{RBs} denote stacks of residual bottleneck blocks~\cite{he2016resnet} shown in the right. \textit{Attn} blocks are attention modules proposed by Cheng~\etal~\cite{cheng2020learned}. \textit{AE} and \textit{AD} are arithmetic en/de-coder, respectively. \textit{TConv} denotes transposed convolution. The blue and red arrows denote the encoding and decoding data flow. The orange ones are shared by both encoding and decoding.}
    \label{fig:arch}
\end{figure*}

Spatial context model and channel-conditional model  eliminate redundancy  along spatial and channel axes. As those dimensions are orthogonal, we assume the redundancy in those dimensions is orthogonal too. Thus, we combine the two models for a better backward-adaptive coding.

Figure~\ref{fig:scctx} shows our space-channel context model (SCCTX).
In the $k$-th unevenly grouped chunk, we apply a spatial context model $g_{\mathrm{sp}}^{(k)}$ to recognise spatial redundancy (eq.~\ref{eq:sp-ctx}). 
It could be an  autoregressive convolution~\cite{minnen2018joint} or its two-pass parallel adaption~\cite{he2021checkerboard}. We introduce  $g_{\mathrm{ch}}$ networks to model the channel-wise context $\mathbf{\Phi}_{\mathrm{ch}}^{(k)}$ (Figure~\ref{fig:unevenly-cc} and eq.~\ref{eq:cc}). The output of spatial and channel branches at the $(k, i)$-th location, $\mathbf{\Phi}_{\mathrm{sp}, i}^{(k)}$ and  $\mathbf{\Phi}_{\mathrm{ch}}^{(k)}$,  will be concatenated with hyperprior representation $\mathbf{\Psi}$ and fed into a location-wise aggregation network to predict the entropy parameters $\mathbf{\Theta}^{(k)}_i = (\bm{\mu}^{(k)}_i, \bm{\sigma}^{(k)}_i)$ for the following en/de-coding of $\hat \vy_i^{(k)}$. Then the just obtained  $\hat \vy_i^{(k)}$ will be used as context to compute $\mathbf{\Phi}_{\mathrm{sp}, (i+1)}^{(k)}$ or $\mathbf{\Phi}_{\mathrm{ch}}^{(k+1)}$, till en/decoding the entire $\hat \vy$.

As shown in Table~\ref{tab:backward-adaption-schemes}, by default we use the parallel checkerboard~\cite{he2021checkerboard} model as the spatial context for SCCTX, which is only applied inside each channel chunk to be more efficient. 

\section{ELIC: efficient learned image compression with scalable residual nonlinearity}

\subsection{Stacking residual blocks for nonlinearity}

For a long period, generalized divisive normalization (GDN) is one of the most frequently used techniques in learned image compression~\cite{balle2018variational, lee2019hybrid, cheng2020learned, xie2021inn, guo2021causal}. It introduces point-wise nonlinearity  to the model~\cite{balle2018efficient}, which  aggregates information along the channel axis and scales feature vectors at each location. Different from linear affine normalization techniques like batch normalization or layer normalization, this nonlinear GDN performs more similarly to point-wise attention mechanism. Thus, we propose to investigate other nonlinear transform layers as alternatives of GDN. Note that this is different from existing investigations that view GDN as activation function~\cite{cheng2019deep, cheng2020learned, kirmemis2020shrinkage, duan2022study-on-gdn}. 

We replace the GDN/IGDN layers with stacks of residual bottleneck blocks~\cite{he2016resnet}. Earlier works also investigate pure convolution networks for visual compression~\cite{cheng2018deep, chen2017deepcoder, cheng2019deep, chen2021nlb-3d} and here we revisit it from the layer-level perspective. We observe that the performance further improves when the number of stacked blocks increases, because of the enhanced nonlinearity. Thus, a network with strong enough nonlinearity can express the intermediate features better for rate-distortion trade-off, even without GDN layers. A similar structure is also investigated by Chen~\etal~\cite{chen2021nlb-3d}, as a part of non-local attention module. We experimentally prove that, even without attention mechanism, the RD performance can still be improved by simply stacking the residual blocks.

Stacking residual blocks allows us to better conduct scalable model profiling. It is also expected to benefit from modern training and boosting techniques like network architecture search~\cite{liu2018darts, yu2020bignas} and loss function search~\cite{li2019amlfs, sun2021hlic}, though they are out of the bound of this work. Residual block is also easier to be extended for dynamic or slimmable inference, while GDN requires special handling~\cite{yang2021slimmable}.

\subsection{Architecture of ELIC}

We summarize the above-mentioned techniques in our proposed model named ELIC (see Figure~\ref{fig:arch}), to build an efficient learned image compression model with strong compression performance. We use the proposed SCCTX module to predict the entropy parameters $\mathbf{\Theta} = (\bm{\mu}, \bm{\sigma})$ of a mean-scale Gaussian entropy model~\cite{minnen2018joint}. Using this more powerful backward-adaptive coding allows adopting lighter transform networks compared with recent prior works~\cite{cheng2020learned, xie2021inn, guo2021causal, gao2021attn-back-proj-iccv}. In the main transforms, we simply adopt stride-2 convolutions to de/in-crease the feature map sizes, following earlier settings~\cite{balle2018variational, minnen2018joint}.

\section{Quickly decoding thumbnail-preview}

\begin{figure}[t]
    \centering
    \includegraphics[width=1.0\linewidth]{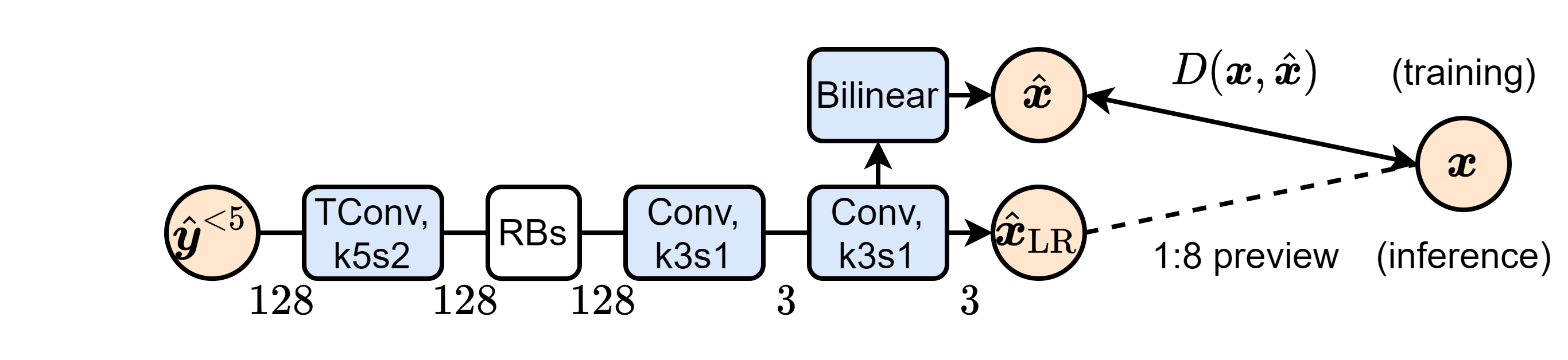}
    \caption{Structure of the proposed thumbnail synthesizer. \textit{Bilinear} module denotes three bilinear upsampling layers by factor 2.}
    \label{fig:diagram-thumbnail-decoder}
\end{figure}

The major bottleneck of decoding process is the synthesis inference, which runs a heavy network to reconstruct the full-resolution image. When applying learned image compression, however, we do not always want to decode the full-resolution image. For instance, when looking through images saved on a server, we require the decoder to quickly generate thousands of thumbnail-preview images which have much lower resolution but keep the structure and semantics of the original images. Another case is progressive decoding preview, investigated in prior literature~\cite{minnen2020channel}. When the image is progressively and partially decoded step by step, frequently invoking the heavy synthesizer will critically slow down the overall decoding process.  On these occasions, image quality is far less important than decoding speed. Thus, directly decoding the full-resolution images using the heavy synthesizer is impractical.

We propose to train an additional tiny network, called thumbnail synthesizer, to reconstruct low-resolution images as thumbnail-preview. When adopting SCCTX, most semantic information is compacted in the earlier decoded channels. Hence we propose to generate the preview image only from the first 4 chunks (\ie the first 128 channels). 

The structure of our proposed thumbnail synthesizer is shown in Figure~\ref{fig:diagram-thumbnail-decoder}. After training the main models, we freeze all the learned parameters and change the main synthesizer to initialized thumbnail synthesizer. Then we restart the distortion optimization to train the model.

As the proposed thumbnail synthesizer is extremely light, its decoding only requires a few microseconds (w.r.t. $768\times 512$ images). Compared with obtaining the preview image by down-sampling from the entirely reconstructed full-resolution image, using the proposed model to obtain the thumbnail-preview images is much more efficient.

\section{Experiments}

\subsection{Settings}

We train the models on the largest 8000 images picked from ImageNet~\cite{deng2009imagenet} dataset. A noising-downsampling preprocessing is conducted following prior works~\cite{balle2018variational, he2021checkerboard}.  We use Kodak~\cite{kodak} and CLIC Professional~\cite{clic2020dataset} for evaluation.

The training settings are accordingly sketched from existing literature~\cite{balle2018variational, minnen2018joint, cheng2020learned, he2021checkerboard}.
For each architecture, we train models with various $\lambda$ standing for different quality presets.  We set $\lambda = \{4, 8, 16, 32, 75, 150, 300, 450\}\times 10^{-4}$ for each model when optimizing MSE. Empirically, on Kodak, models trained with these $\lambda$ values will achieve average bits-per-pixel (BPP) ranging from $0.04$ to $1.0$. We set the number of channels $N=192$ and $M=320$ for all models. We train each model with an Adam optimizer with $\beta_1 = 0.9, \beta_2 = 0.999$. We set initial learning-rate to $10^{-4}$, batch-size to 16 and train each model for 2000 epochs (1M iterations, for ablation studies) or 4000 epochs (2M iterations, to finetune the reported ELIC models), and then decay the learning-rate to $10^{-5}$ for another 100-epoch training.

We use mixed quantization estimator to train channel-conditional models, following Minnen~\etal~\cite{minnen2020channel}. It helps optimize the single Gaussian mean-scale entropy model~\cite{minnen2018joint}, making it comparable with mixture models like GMM~\cite{lee2019hybrid, liu2020edic, cheng2020learned, guo2021causal}. Following prior works~\cite{minnen2018joint, he2021checkerboard} and community discussion\footnote{\url{https://groups.google.com/g/tensorflow-compression/c/LQtTAo6l26U/m/mxP-VWPdAgAJ}}, we encode each $\left \lceil y-\mu \right \rfloor$ to the bitstream instead of $\left \lceil y \right \rfloor$ and restore the coding-symbol as $\left \lceil y-\mu \right \rfloor + \mu$, which can further benefit the single Gaussian entropy model. When using the autoregressive context model with the two above-mentioned strategies together, the one-pass encoding~\cite{he2021checkerboard} doesn't simply work for training because of the inconsistency in synthesizer input ($\left \lceil y-\mu \right \rfloor + \mu$ instead of $\left \lceil y \right \rfloor$). Therefore, we adopt a two-stage training. In the beginning 2000/3800 epochs, we sample uniform noise to estimate $\hat y$ input to SCCTX and use $\mathrm{STE}(y)$ as the input of synthesizer. Thus, the faster one-pass encoding can be adopted. Then, we fed $\mathrm{STE}(y-\mu) + \mu$ to both the synthesizer and SCCTX in the remained epochs.

To obtain a lightweight model, we do not adopt the LRP network~\cite{minnen2020channel}, and the low-rate models ($\lambda \le 0.0075$) perform worse if trained with target $\lambda$ values from  scratch. Inspired by prior work~\cite{minnen2020channel}, we train these models with $\lambda=0.015$ at the beginning, and adapt them using target $\lambda$ values after decaying the learning-rate.

Please refer to the supplementary material for more detailed experiment settings, including running environment and raw data accessing.

\begin{figure}[tb]
    \centering
    \subfloat{
    \includegraphics[width=0.98\linewidth]{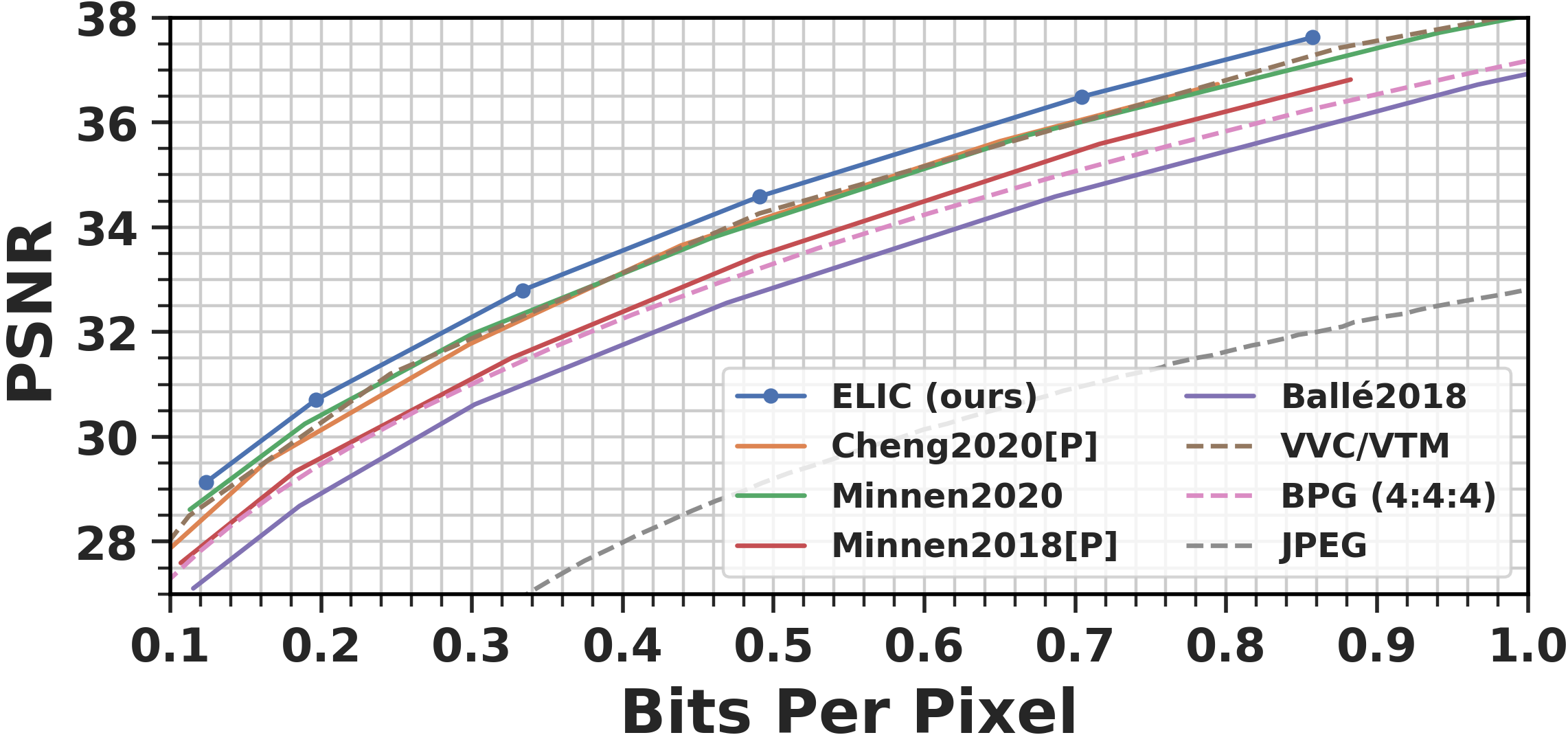}
    }
    \\
    \subfloat{
    \includegraphics[width=0.98\linewidth]{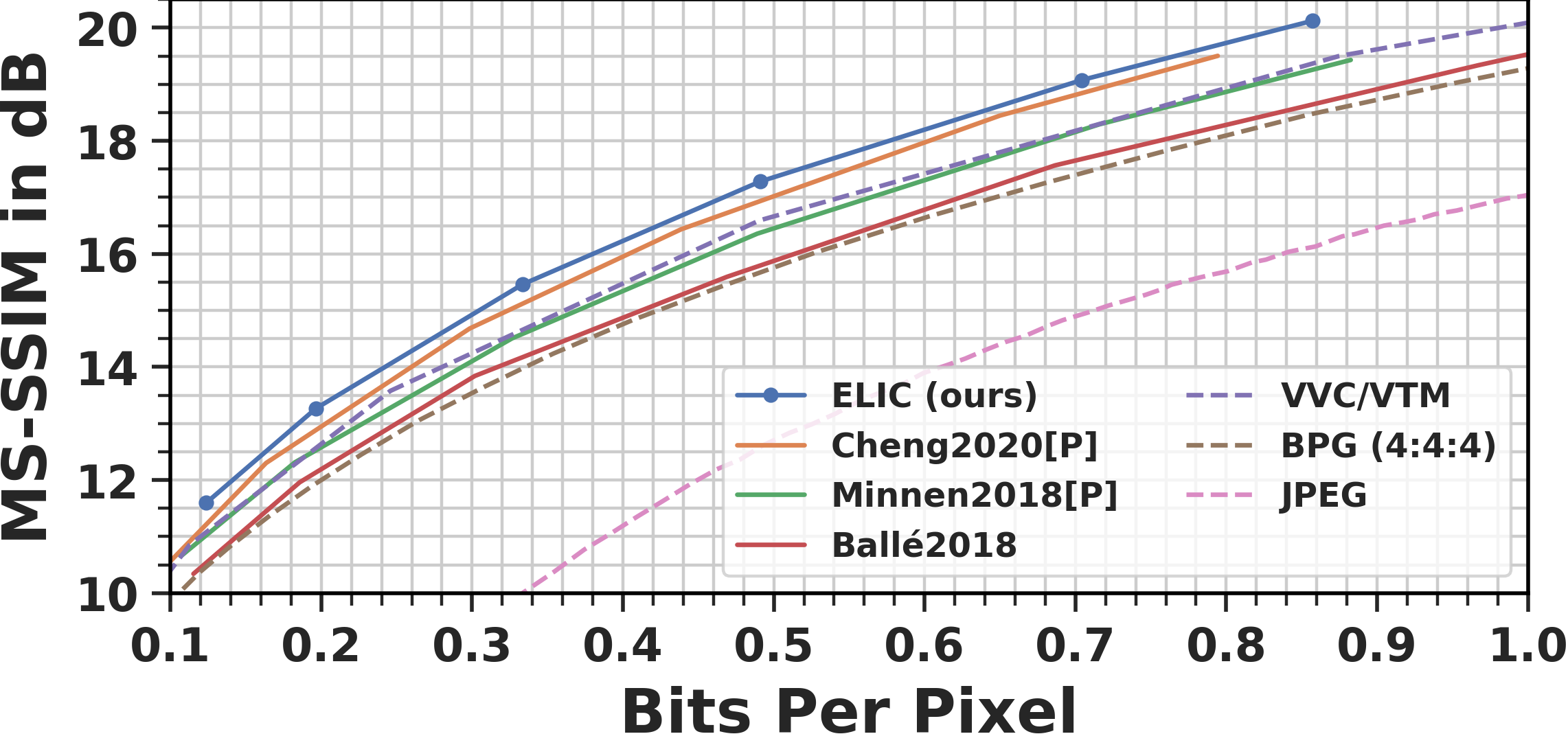}
    }
    \caption{Rate-distortion curves of various image compression approaches. The results are evaluated on Kodak. All shown learned models are optimized for minimizing MSE.}
    \label{fig:comp-psnr}
\end{figure}

\newcommand{\graycell}{\cellcolor[gray]{0.85}}
\begin{table*}[htb]
    \centering
    \scalebox{0.9}{
    \begin{tabular}{lcrrr|rrrrr}
    \toprule
    \multirow{2}{*}{\textbf{Model}} & \multirow{2}{*}{\textbf{Sp. Context}} &
          \multirow{2}{*}{\textbf{BD-Rate (\%)}} &  \multicolumn{7}{c}{\textbf{Inference Latency (ms)}} \\
         &&\textbf{}& \textbf{Enc. Tot.} & \textbf{Dec. Tot.} & \textbf{YEnc.}& \textbf{YDec.} & \textbf{ZEnc.}& \textbf{ZDec.}  & \textbf{Param.}  \\
         \midrule
         ELIC (ours) & [P] & \textbf{-7.88} & 42.44 & 49.16 & 31.16 & 38.33 & 1.18 & 1.59 & 10.10\\
         ELIC-sm (ours) & [P] & -1.07 &  22.64 & 27.80 & 12.55 & 17.37 & 1.07 & 1.41 &  9.02 \\
         Minnen2020~\cite{minnen2020channel} & - & 1.11 & 60.76 & 65.75 & 10.60 & 15.41 & 1.36 & 1.54 & 48.80\\
         Cheng2020[P]~\cite{he2021checkerboard}& [P] & 3.89 & 45.43 &53.01 & 36.84& 44.20& 0.93 & 1.15 & 7.66\\
         Minnen2018[P]~\cite{he2021checkerboard} & [P] & 20.00 &18.84& 24.52 & 11.63 & 17.32 &  1.47&  1.46&  5.74 \\
         Ball\'e2018~\cite{balle2018variational} &- & 40.85 & 
         13.07 & 17.24 & 11.90& 16.14 & 1.17& 1.10 & -
         \\
         \midrule
         Gao2021~\cite{gao2021attn-back-proj-iccv} & [S] & \textbf{-10.94} & / & / & / & / & / & / & / \\
         Guo2021~\cite{guo2021causal} & [S] & -7.02 & 190.41 &  $>10^{3}$ &  154.08 & 431.68 & 3.67 & 1.61 &  $>10^{3}$  \\
         Xie2021~\cite{xie2021inn} & [S] & -0.54 & 61.20 & $>10^{3}$ & 57.96 & 162.18 & 1.28 & 1.34 &  $>10^{3}$ \\
         Cheng2020~\cite{cheng2020learned} & [S] & 3.35& 41.42 &  $>10^{3}$ & 36.99& 47.35 & 0.96 &1.12 & $>10^{3}$\\
         Minnen2018~\cite{minnen2018joint} & [S] & 14.92 & 18.14 &   $>10^{3}$ &  12.75& 17.34 & 1.69 & 1.60 & $>10^{3}$ \\
         \midrule
         VVC (YUV 444)~\cite{vtm2019} & - &0.00  & - & - & - & - & - & - & - \\
         BPG~\cite{bellard2015bpg} & - & 27.21 & - & - & - & - & - & - & -\\
         JPEG~\cite{JPEG-ITU1992Information} & - & 266.32 & - & - & - & - & - & - & - \\
         \bottomrule
    \end{tabular}
    }
    \caption{RD and inference time of learned image compression models. The BD-Rate data is calculated relative to VVC (YUV 444) from PSNR-BPP curve on Kodak. The \textbf{Param.} column denotes inference latency of entropy parameter calculation during decoding (\eg the running time of SCCTX in our ELIC model). \textbf{Enc. Tot.} and \textbf{Dec. Tot.} denote total network latency for encoding and decoding respectively. Models marked with [P] adopt parallel checkerboard context model and [S] denotes serial context model. We have not reproduced the very recent Gao2021 model, yet its speed reported by the authors~\cite{gao2021attn-back-proj-iccv} is slower than the referred Cheng2020 model (more than 1s/image).}
    \label{tab:comp-full-table}
\end{table*}

\subsection{Quantitative results}

To compare the RD performance of various models, we present BD-rate~\cite{bjontegaard2001bdbr} computed from PSNR-BPP curves as the quantitative metric. The anchor RD performance is set to VVC or BPG accordingly (with an anchor BD-rate $0\%$). In the supplementary material we also show BD-rate results calculated on MS-SSIM.

\subsubsection{Comparison of coding performance and speed}

We compare the RD performance and coding speed of ELIC with exiting learning based approaches,
as shown in Table~\ref{tab:comp-full-table}. Apparently, our ELIC achieves remarkable improvement on compression quality and speed among learned image compression approaches. We also report BD-rates of manual-designed image compression approaches, but do not compare the speed of them because they cannot run on GPU.  ELIC outperforms VVC (in YUV 4:4:4 colorspace) on RD performance regarding PSNR (and also MS-SSIM, see Figure~\ref{fig:comp-psnr}).
Without expensive RDO searching process, learned compression models including ELIC can achieve encoding speed as fast as decoding speed, which is potentially helpful to conduct high-quality real-time visual data transmission. This can be an advantage of learning based approaches, since currently best conventional image coders like VVC spend much longer time in encoding than decoding\footnote{ \url{https://github.com/InterDigitalInc/CompressAI/blob/v1.1.8/results/kodak/vtm.json} by CompressAI~\cite{begaint2020compressai}}. 

\begin{table}[b]
    \centering
    \begin{tabular}{clrr}
    \toprule
    \multirow{2}{*}{\textbf{Data}} & \textbf{Grouping} & \multirow{2}{*}{\textbf{BD-Rate}}  & \textbf{Param. } \\
    &\textbf{Scheme} && \textbf{Latency}\\
         \midrule
         \multirow{5}{*}{Kodak} 
          &even (10-slice) & -3.81 &  13.46 \\
          &+ spatial[P] & -5.02 &  19.41\\
          &uneven & -2.78 & \textbf{6.45}\\
          &+ spatial[P] & \textbf{-5.63} & 10.10 \\
          &+ spatial[S] & {\color{gray} -6.99} & {\color{gray} $>10^3$} \\
         \midrule
         \multirow{5}{*}{CLIC-P} 
          &even (10-slice) & -12.42 & 73.59 \\
          &+ spatial[P] & -14.11 & 85.44 \\
          &uneven & -11.54 & \textbf{31.88} \\
          &+ spatial[P] & \textbf{-14.51} & 46.06 \\
          &+ spatial[S] & {\color{gray} -15.51} & {\color{gray} $>10^3$} \\
         \midrule
         VVC &-& 0.00 & - \\
         \bottomrule
    \end{tabular}
    \caption{Performance of different grouping schemes. Since image resolutions in CLIC Professional vary, we report the latency repeatedly tested on its largest $2048\times 1890$ image.}
    \label{tab:comp-grouping}
\end{table}

We present a slim architecture ELIC-sm, which is adapted from ELIC by removing attention modules and using fewer res-blocks (RB$\times 1$). It also achieves a remarkable RD performance while significantly reducing the latency.

To further present our approach, we highlight Figure~\ref{fig:comp-bdbr-speed} and Figure~\ref{fig:comp-psnr}. Figure~\ref{fig:comp-bdbr-speed} shows the RD-latency relationship of various approaches. It can be seen from the figure that our ELIC model achieves state-of-the-art performance regarding Pareto optimum. We also draw all tested learned image compression approaches which can decode the Kodak image in 100 microseconds in Figure~\ref{fig:comp-psnr}, which further indicates the superiority of ELIC. For completeness, we present more RD curves and results tested on Kodak and other datasets in the supplementary material.

\subsubsection{Ablation study}

\textbf{Influence of uneven grouping and spatial adaption in SCCTX.} We evaluate the proposed uneven grouping model and compare it with the previous even grouping approach~\cite{minnen2020channel}. For a fair comparison, we train a line of models with the same architectures for main and hyper auto-encoders as ELIC. The only difference between them is the grouping models for backward-adaptive coding. Table~\ref{tab:comp-grouping} shows the results evaluated on Kodak and CLIC Professional. With the proposed uneven grouping, the latency of adaptive entropy estimation decreases by half.
Introducing the spatial context model to both even and uneven models gain RD promotion, with uneven+spatial[P] running speed still faster than the 10-slice even grouping models. 

\begin{table}[tb]
    \centering
    \begin{tabular}{lrrrr}
    \toprule
    \multirow{2}{*}{\textbf{Model}} &
         \multirow{2}{*}{\textbf{Layer}}   & \multirow{2}{*}{\textbf{BD-Rate}} &  \multicolumn{2}{c}{\textbf{Latency}} \\
         &&& \textbf{Enc.}& \textbf{Dec.} \\
         \midrule
         Ball\'e2018 & GDN & 8.23 & 12.27 & 17.30\\
         Ball\'e2018 & RB$\times 1$ & 5.68 & 12.45 & 17.51\\
         Ball\'e2018 & PWRB$\times 3$ & 4.59 & 18.15 & 22.84\\
         Ball\'e2018 & RB$\times 3$ & \textbf{-2.94} & 23.00 &27.91\\
         \midrule
         Minnen2018 & GDN & -5.04 & 12.75 & 17.34\\
         Minnen2018 & RB$\times 1$ & -6.34 & 12.90 &  17.44\\
         Minnen2018 & PWRB$\times 3$ & -11.14 & 18.10 & 22.78\\
         Minnen2018 & RB$\times 3$ & \textbf{-13.56} & 23.01 & 28.21\\
         \midrule
         BPG & - & 0.00 & - & - \\ 
         \bottomrule
    \end{tabular}
    \caption{Comparison of different nonlinear layers, evaluated on Kodak. The columns marked as \textbf{Latency Enc./Dec.} denote inference time of main analysis/synthesis. \textit{RB}$\times n$ denotes a stack of  $n$ residual bottleneck blocks~\cite{he2016resnet} (the right in Figure~\ref{fig:arch}). \textit{PWRB} is point-wise residual block which removes the $3\times 3$ convolution.}
    \label{tab:comp-nonlinear-layers}
\end{table}

\textbf{Residual blocks versus GDN (Table~\ref{tab:comp-nonlinear-layers}).} After replacing all GDN/IGDN layers with residual bottleneck blocks, the RD performance improves on both Ball\'e2018~\cite{balle2018variational} and Minnen2018~\cite{minnen2018joint} baselines while the inference latency is still on par with GDN version. When stacking more RB blocks as nonlinear transform, the BD-rates even reduce more, which proves the scalability. We also try to stack GDN layers but the training becomes very unstable and eventually fails.

Since the residual bottleneck module has a larger receptive field than GDN, we also evaluate a point-wise version of it (\textit{PWRB}) to figure out the influence of a larger receptive field. It still improves BD-rates on both baselines. Note that residual blocks with larger receptive fields improve more on models without spatial context (Ball\'e2018), implying that nonlinear transform with larger receptive fields is also helpful to remove spatial redundancy.

\subsection{Qualitative results}

\textbf{Full reconstruction.} We present the high-resolution images coded by ELIC and prior approaches in the supplementary material to show the reconstruction performance.

\textbf{Thumbnail-preview.} See Figure~\ref{fig:thumbnail-preview}, where we present the thumbnail-preview of Kodak images generated from our thumbnail synthesizer. Since the thumbnail-preview images are in low-resolution by design, the relative low reconstruction quality (PSNR $=23.02$ dB evaluated on full-resolution Kodak) will not do harm to subjective feeling, as major semantic information remains and the artifacts are suppressed by image downsampling. 
The inference of thumbnail synthesizer takes about 3 microseconds, which is over 12 times faster than invoking the full synthesizer of ELIC. 

\newcommand{\thumbimgwidth}{0.14\linewidth}

\begin{figure}[tb]
    \centering
    \subfloat{
    \includegraphics[width=\thumbimgwidth]{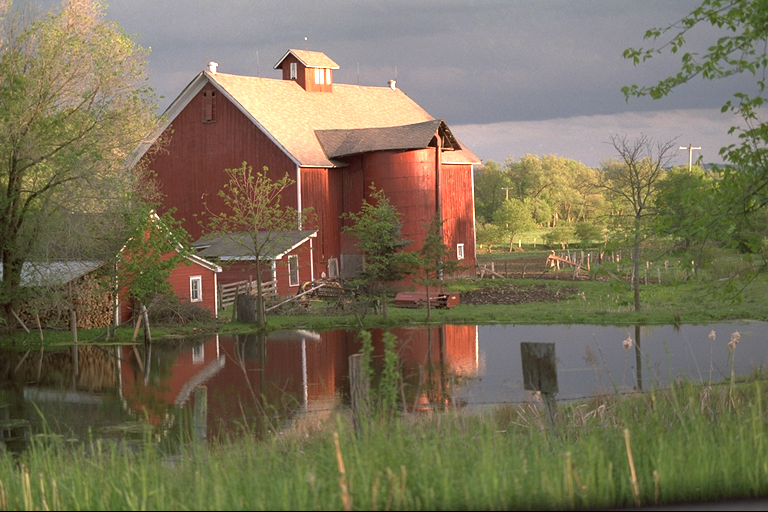}
    }
    \subfloat{
    \includegraphics[width=\thumbimgwidth]{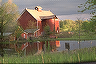}
    }
    \subfloat{
    \includegraphics[width=\thumbimgwidth]{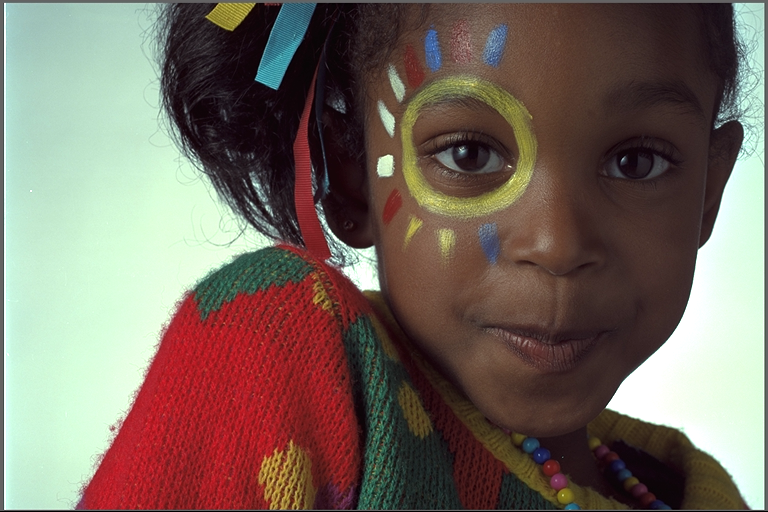}
    }
    \subfloat{
    \includegraphics[width=\thumbimgwidth]{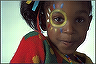}
    }
    \subfloat{
    \includegraphics[width=\thumbimgwidth]{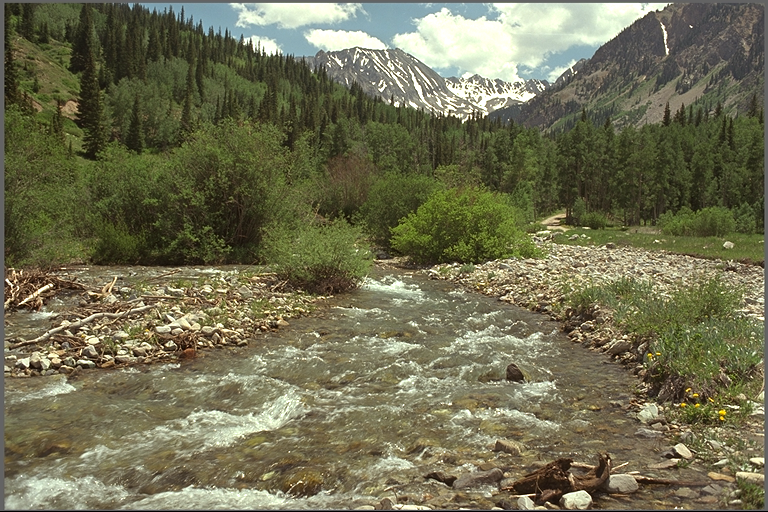}
    }
    \subfloat{
    \includegraphics[width=\thumbimgwidth]{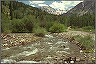}
    }
    \\
    \subfloat{
    \includegraphics[width=\thumbimgwidth]{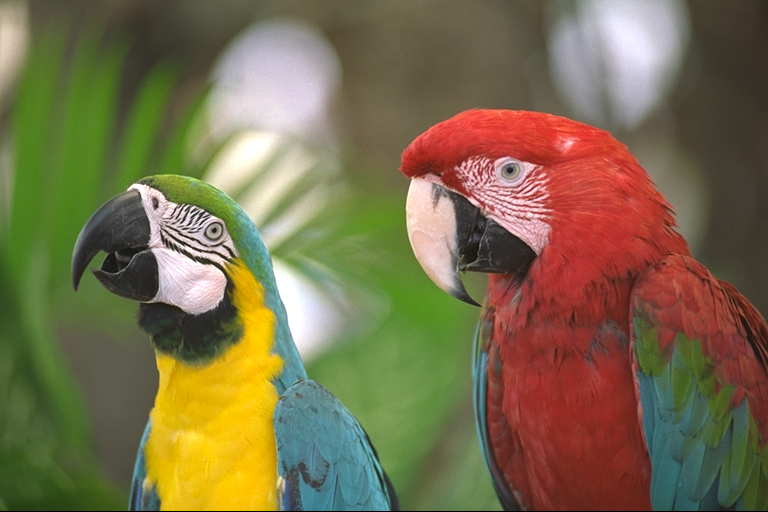}
    }
    \subfloat{
    \includegraphics[width=\thumbimgwidth]{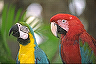}
    }
    \subfloat{
    \includegraphics[width=\thumbimgwidth]{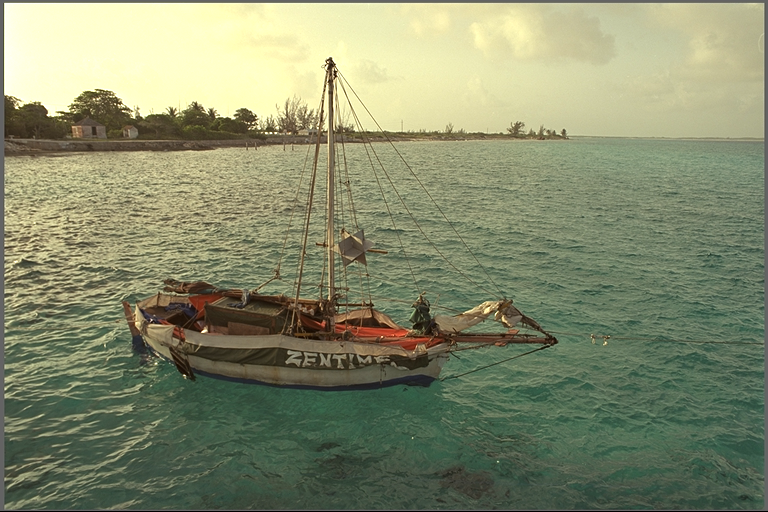}
    }
    \subfloat{
    \includegraphics[width=\thumbimgwidth]{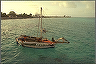}
    }
    \subfloat{
    \includegraphics[width=\thumbimgwidth]{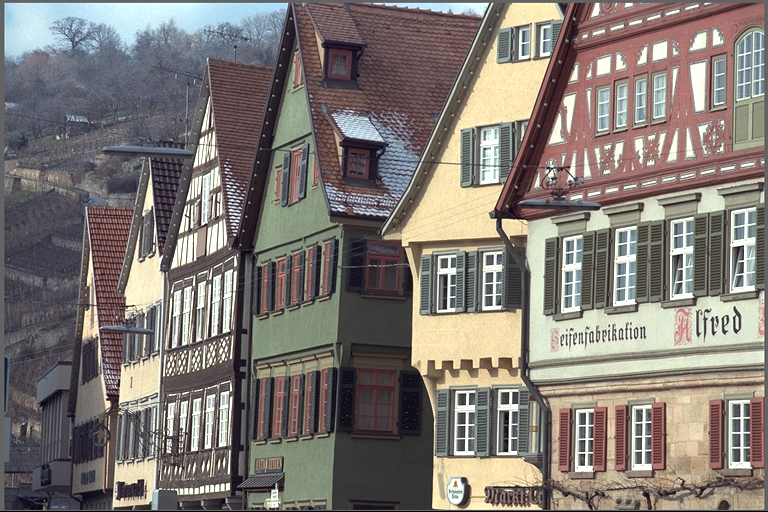}
    }
    \subfloat{
    \includegraphics[width=\thumbimgwidth]{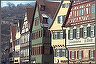}
    }
    \caption{Quickly decoded thumbnail-preview images. For each pair of images, the left one is the downsampled ground-truth and the right is corresponding thumbnail-preview.}
    \label{fig:thumbnail-preview}
\end{figure}

\textbf{Progressive decoding} is an extended application of channel-conditional models~\cite{minnen2020channel}. We also present the progressive decoding results of ELIC.  Different from Minnen~\etal~\cite{minnen2020channel}, at the $k$-th step, we directly fill the non-decoded channel chunks $\hat \vy^{(k+1)}, \dots, \hat \vy^{(5)}$ with 0 before feeding them to the synthesizer. Therefore, extra calculation predicting non-decoded symbols is avoided. See Figure~\ref{fig:progressive-dec}. With the beginning 16 channels, the structural information can be already reconstructed. The following channel groups further provide chrominance and high-frequency information. Since the progressive decoding is usually adopted for preview, we use the thumbnail synthesizer to quickly reconstruct the partially decoded ($k\le 4$) images.

\newcommand{\progressiveimgwidth}{0.3\linewidth}
\begin{figure}[t]
    \centering
    \subfloat[$k=1$]{
    \includegraphics[width=\progressiveimgwidth]{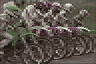}
    }
    \subfloat[$k=2$]{
    \includegraphics[width=\progressiveimgwidth]{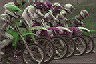}
    }
    \subfloat[$k=3$]{
    \includegraphics[width=\progressiveimgwidth]{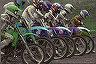}
    }
    \\
    \subfloat[$k=4$]{
    \includegraphics[width=\progressiveimgwidth]{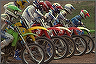}
    }
    \subfloat[$k=5$]{
    \includegraphics[width=\progressiveimgwidth]{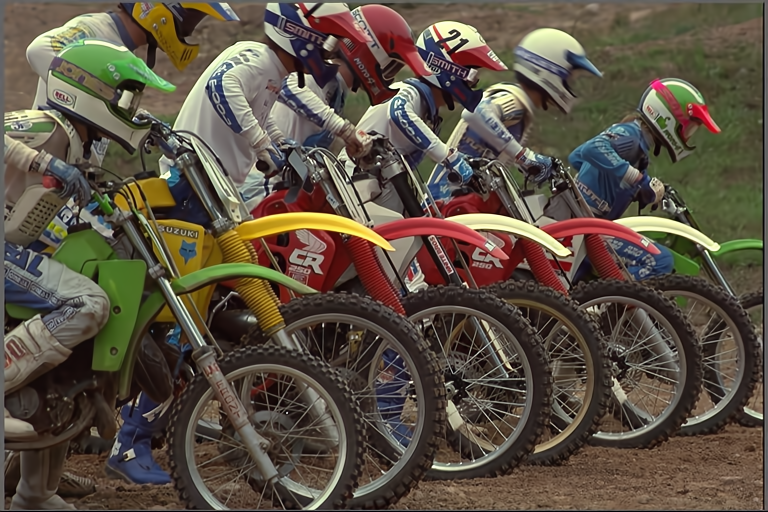}
    }
    \subfloat[ground truth]{
    \includegraphics[width=\progressiveimgwidth]{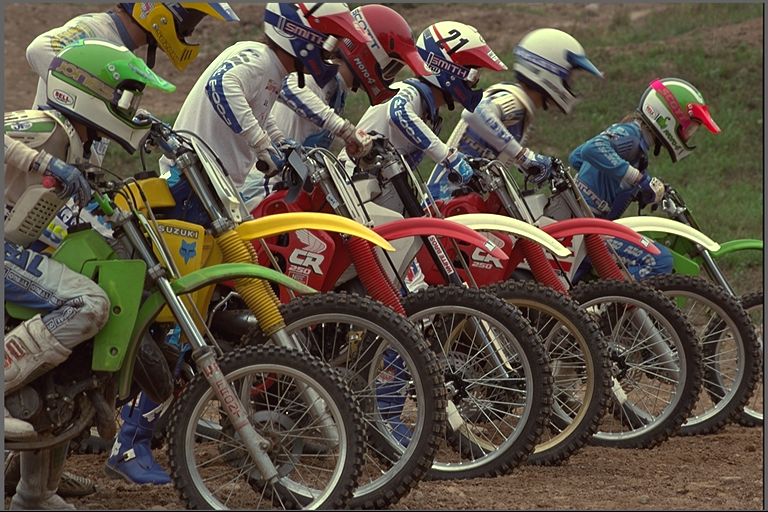}
    }
    \caption{Progressive decoding. Each image is reconstructed from the first $k$ groups. When $k=5$, the image is fully decoded.}
    \label{fig:progressive-dec}
\end{figure}

\section{Discussion and conclusion}
With the proposed SCCTX and the residual transform networks, we obtain state-of-the-art model ELIC, which better balances compression ability and running speed. Furthermore, we propose to train a thumbnail network for preview decoding, which also improves the utility. 
In the future, we will further investigate the information compaction phenomenon for improving the architecture.

Note that VVC is majorly designed for YUV 4:2:0 colorspace instead of YUV 4:4:4, as the former better reflects the sensitivity of human perception. We will also delve into models with both objective and subjective image quality remarkable, following exiting literature~\cite{egilmez2021transform, mentzer2020hific, blau2019rethinking-rdp-tradeoff, patel2021saliency}.

\balance
{\small
\bibliographystyle{ieee_fullname}
\bibliography{ms}
}

\end{document}


\title{Supplementary Material \\ \vspace{10pt} \small{ELIC: Efficient Learned Image Compression with  Unevenly Grouped Space-Channel Contextual Adaptive Coding}\vspace{-30pt}}

\author{}
\maketitle

\section{Detailed network architecture}

\subsection{Architecture of transform networks}

\begin{table}[htb]
    \centering
    \scalebox{1}{
    \begin{tabular}{c|c}
    \toprule
         Analyzer $g_a$ & Synthesizer $g_s$\\
    \midrule
        in: $3$-channel image & in: $M$-channel symbols \\
    \midrule
         Conv $5\times 5$, s2, $N$ & Attention \\
         ResBottleneck$\times 3$ & TConv $5\times 5$, s2, $N$\\
         Conv $5\times 5$, s2, $N$ & ResBottleneck$\times 3$\\
         ResBottleneck$\times 3$ & TConv $5\times 5$, s2, $N$\\
         Attention & Attention\\
         Conv $5\times 5$, s2, $N$ & ResBottleneck$\times 3$\\
         ResBottleneck$\times 3$ & TConv $5\times 5$, s2, $N$\\
         Conv $5\times 5$, s2, $M$ & ResBottleneck$\times 3$ \\
         Attention & TConv $5\times 5$, s2, $3$\\
    \bottomrule
    \end{tabular}
    }
    \caption{Architecture of ELIC main transform networks.}
    \label{tab:arc-main-net}
\end{table}

\begin{table}[htb]
    \centering
    \scalebox{1}{
    \begin{tabular}{c|c}
    \toprule
         Analyzer $g_a$ & Synthesizer $g_s$\\
    \midrule
    in: $3$-channel image & in: $M$-channel symbols \\
    \midrule
         Conv $5\times 5$, s2, $N$ &  TConv $5\times 5$, s2, $N$\\
         ResBottleneck$\times 1$ & ResBottleneck$\times 1$\\
         Conv $5\times 5$, s2, $N$ & TConv $5\times 5$, s2, $N$\\
         ResBottleneck$\times 1$ & ResBottleneck$\times 1$\\
         Conv $5\times 5$, s2, $N$ & TConv $5\times 5$, s2, $N$\\
         ResBottleneck$\times 1$ & ResBottleneck$\times 1$ \\
         Conv $5\times 5$, s2, $M$ & TConv $5\times 5$, s2, $3$ \\
    \bottomrule
    \end{tabular}
    }
    \caption{Architecture of ELIC-sm main transform networks.}
    \label{tab:arc-main-net-sm}
\end{table}

As shown in Table~\ref{tab:arc-main-net} and Table~\ref{tab:arc-main-net-sm}, our main networks are adapted from those previously used by Ball\'e~\etal~(2018), Minnen~\etal~(2018) and Minnen~\etal~(2020). We replace GDN/IGDN layers by residual blocks, as mentioned in the main text. We follow previous works to introduce a variable $N$ representing the channel number of intermediate features, and lift the output channel number of analyzer to $M$. Thus, the coding-symbol $\hat \vy$ is an $H \times W \times M$ tensor. 

We adopt residual bottleneck structures as nonlinear transform, which have a dimensional bottleneck which can both provide larger receptive field and keep the volume of calculation acceptable. The channel number of intermediate features before and after $3\times 3$ convolution is $\frac N 2$. We sequentially stack 3 such residual blocks after each down/up-sampling convolution in ELIC (Table~\ref{tab:arc-main-net}), and instead use only one at each position in ELIC-sm (Table~\ref{tab:arc-main-net-sm}). We also use the attention modules adopted by Cheng \etal in the larger ELIC models to further enhance the nonlinearity.

We simply adopt the three-layer hyper analyzer and synthesizer, following previous works (Minnen~\etal, 2018; Minnen~\etal, 2020; He~\etal, 2021). The hyper synthesizer output is an $H\times W\times (2M)$ tensor $\mathbf{\Psi}$ as shown in Figure 7 in the main text.

\subsection{Architecture of SCCTX networks}

Following Minnen~\etal~(2020), we use $5\times 5$ convolutions to analyze cross-channel redundancy. The architecture of $g_{\mathrm{ch}}$ network is frankly sketched from Minnen~\etal~(2020). We follow the suggestions of Minnen~\etal~(2018) and He~\etal~(2021) and use single-layer $5\times 5$ autoregressive/checkerboard-masked convolution as spatial context model $g_\mathrm{sp}$. Both $g_\mathrm{sp}^{(k)}$ and $g_\mathrm{ch}^{(k)}$ output $H\times W \times (2M^{(k)})$ features, where $M^{(k)}$ is the channel number of the $k$-th channel group.

Therefore, we have the concatenated adaptive representation tensor $[\mathbf{\Phi}_{\mathrm{sp}, i}^{(k)}, \mathbf{\Phi}_{\mathrm{ch}}^{(k)}, \mathbf{\Psi}] \in \mathbb{R}^{H\times W\times (4M^{(k)} + 2M)}$. The parameter aggregation network linearly reduces the dimensions to $2M^{(k)}$.

\section{Detailed experimental settings}

We implement, train, and evaluate all learning-based models on PyTorch~1.8.1. We use NVIDIA TITANXP to test both RD performance and inference speed. To test the speeds, we reproduce previously proposed models and evaluate them under the same running conditions for fair comparison. 
Since most of the models adopt reparameterization techniques, we fix the reparameterized weights before testing the speed. We follow a common protocol to test the latency with GPU synchronization. When testing each model, we drop the latency results (we do not drop them when evaluating RD performance) of the first 6 images to get rid of the influence of device warm-up, and average the running time of remained images to get the precious speed results. 

We do not enable the deterministic inference mode (\eg \texttt{torch.backends.cudnn.deterministic}) when testing the model speeds for two reasons. First, we tend to believe that the deterministic issue can be well solved with engineering efforts, such as using integer-only inference. Thus, the deterministic floating-point inference is unnecessary. Second, the deterministic mode extremely slows down the speed of specific operators, like transposed convolutions which are adopted by ELIC and earlier baseline models (Ball\'e \etal and Minnen \etal), making the comparison somewhat unfair.

We obtain the RD results of prior works by asking the authors via emails or accessing released data directly.

\subsection{Visualization explanation}

We visualize the coding-symbols to investigate the information compaction property. The visualization results of the lighthouse image is shown in Figure~2 of the main text. The $\ell$-th gray scale image is the rescaled magnitude $\vf^{\ell}$ of the $\ell$-th symbol channel $\hat \vy ^{(\ell)}$:
\begin{equation}
    \vf^{(\ell)} = \frac {|\hat \vy ^{(\ell)}|} {\max \hat \vy}, \quad \ell = 1,2\dots,M
\end{equation}

Inspired by the concept of energy in the signal processing community, we introduce the term energy to describe the average square value of each symbol channel:
\begin{equation}
    e^{(\ell)} = \frac 1 {HW} \sum_{\hat y_i^{(\ell)} \in \hat \vy^{(\ell)}} {\hat y_i^{(\ell)2}}
\end{equation}
In the main text, Figure~2 presents the channels with the largest energy values. Figure~3-left further shows the logarithmic energy $\log(e)$ of each channel. 

\section{More rate-distortion results}

We also calculate the BD-rate over VVC when adopting MS-SSIM as the distortion metric. The results are shown in Table~\ref{tab:sup-comp-full-table}. Several baselines are skipped since the original authors do not provide the MS-SSIM results evaluated on the MSE-optimized models.

\begin{table}[htb]
    \centering
    \scalebox{1}{
    \begin{tabular}{lrr}
    \toprule
    \multirow{2}{*}{\textbf{Model}} &
          \multirow{1}{*}{\textbf{BD-Rate (\%)}} & \multirow{1}{*}{\textbf{BD-Rate (\%)}}  \\
         &\textbf{(PSNR)}&  \textbf{(MS-SSIM in dB)} \\
         \midrule
         ELIC (ours) & \textbf{-7.88} & \textbf{-12.73} \\
         ELIC-sm (ours) & -1.07 & -5.59 \\
         Minnen2020 & 1.11 & -\\
         Cheng2020[P] & 3.89 & -7.19 \\
         Minnen2018[P] & 20.00 & 4.23 \\
         Ball\'e2018 & 40.85 & 16.41 \\
         \midrule
         Gao2021 & -10.94 & - \\
         Guo2021 & -7.02 & - \\
         Wu2021~[43] & -5.71 & - \\
         Xie2021 & -0.54 & -7.82 \\
         Cheng2020 &3.35& -3.17\\
         Minnen2018 & 14.92 & -0.68\\
         \midrule
         VVC &0.00 & 0.00\\
         \bottomrule
    \end{tabular}
    }
    \caption{BD-rates over VVC for learned image compression models. The BD-Rate data is calculated from rate-distortion curves over data points with BPP $< 1$ on Kodak. We present results of two distortion metrics, PSNR and MS-SSIM in dB. Models marked with [P] adopt parallel checkerboard context model and [S] denotes serial context model.
    }
    \label{tab:sup-comp-full-table}
\end{table}

\begin{figure*}
    \centering
    \includegraphics[width=0.9\linewidth]{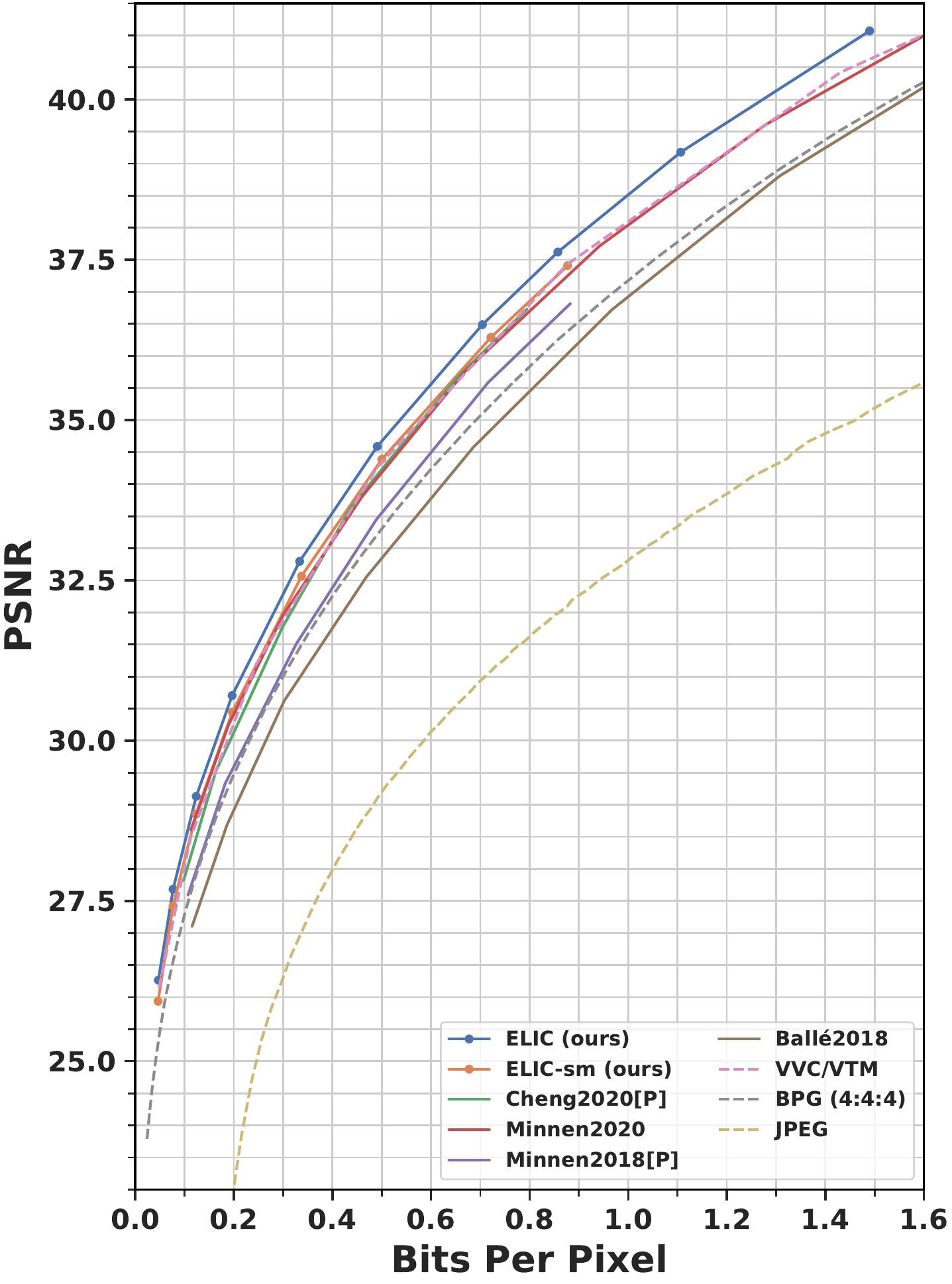}
    \caption{\textbf{PSNR-BPP curve on Kodak.} All models are optimized for minimizing MSE. All presented learning-based models can decode the $512\times 768$ Kodak image in 100 microseconds.}
    \label{fig:comp-psnr-large}
\end{figure*}

\begin{figure*}
    \centering
    \includegraphics[width=0.9\linewidth]{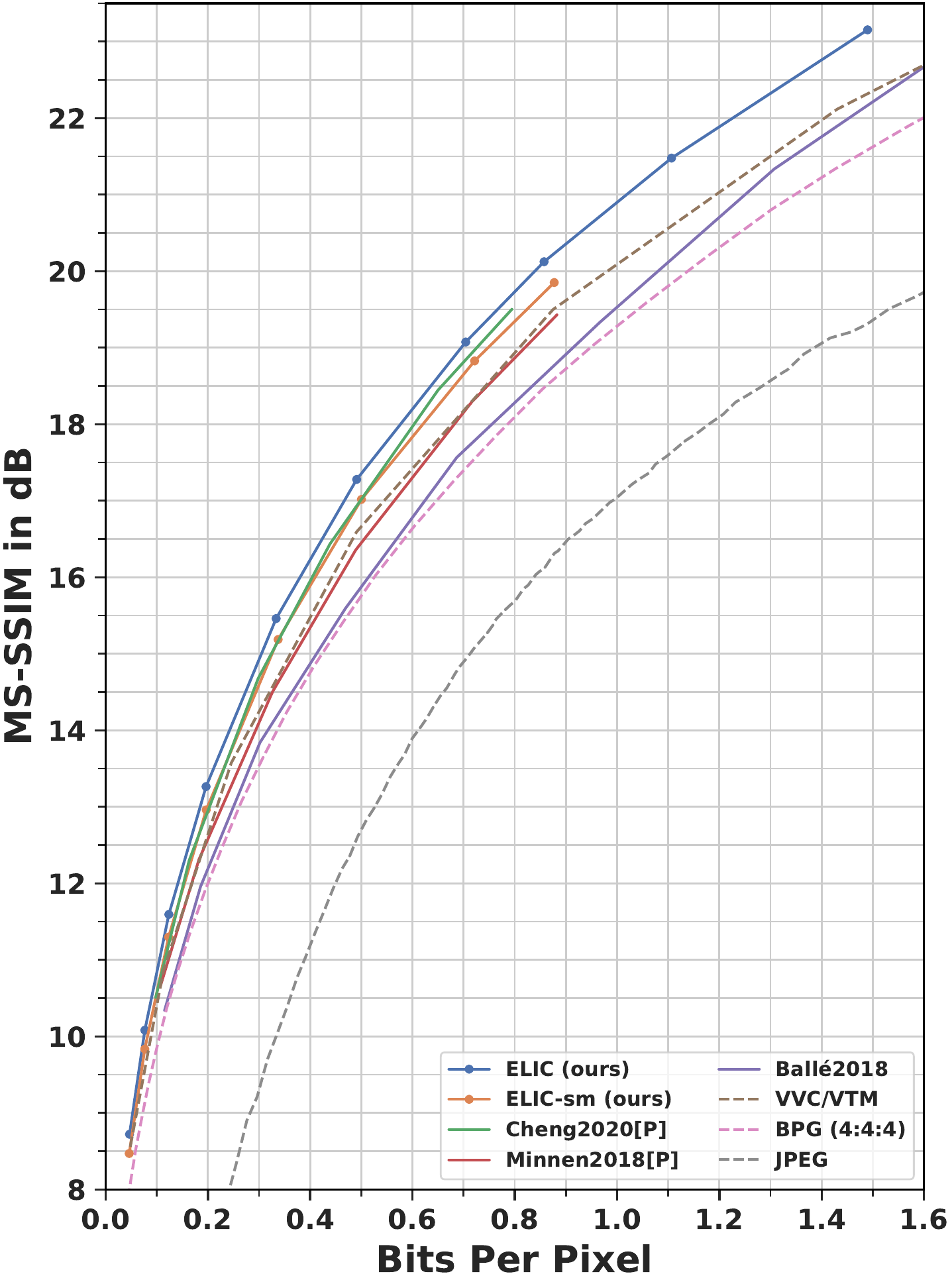}
    \caption{\textbf{MS-SSIM RD curve on Kodak.} All models are optimized for minimizing MSE. All presented learning-based models can decode the $512\times 768$ Kodak image in 100 microseconds.}
    \label{fig:comp-msssim-large}
\end{figure*}

\begin{figure*}
    \centering
    \includegraphics[width=0.9\linewidth]{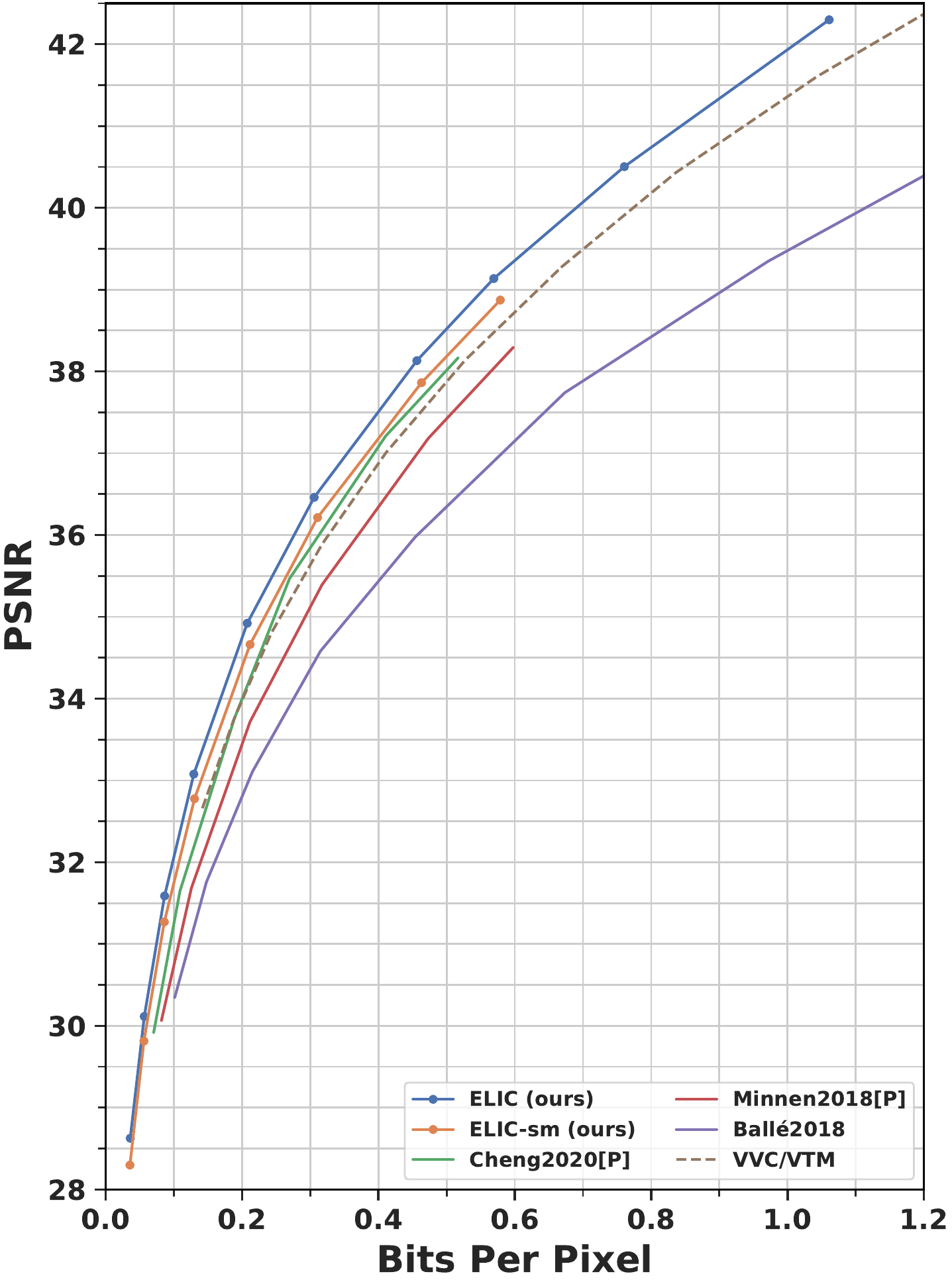}
    \caption{\textbf{PSNR-BPP curve on CLIC-Professional.} All models are optimized for minimizing MSE. All presented learning-based models can decode the $512\times 768$ Kodak image in 100 microseconds.}
    \label{fig:comp-psnr-large-clicp}
\end{figure*}

For completeness, we evaluate our ELIC model at larger BPP range (from 1.0 to 1.5 on Kodak). We train these models with $\lambda=\{0.08, 0.16\}$. The high-rate models still perform well. 
We draw the results on larger figures (Figure~\ref{fig:comp-psnr-large} for PSNR and Figure~\ref{fig:comp-msssim-large} for MS-SSIM) to more clearly show the superiority of the proposed approach. We also evaluate the models on CLIC-Professional and report Figure~\ref{fig:comp-psnr-large-clicp}.

\begin{figure*}
    \centering
    \includegraphics[width=0.9\linewidth]{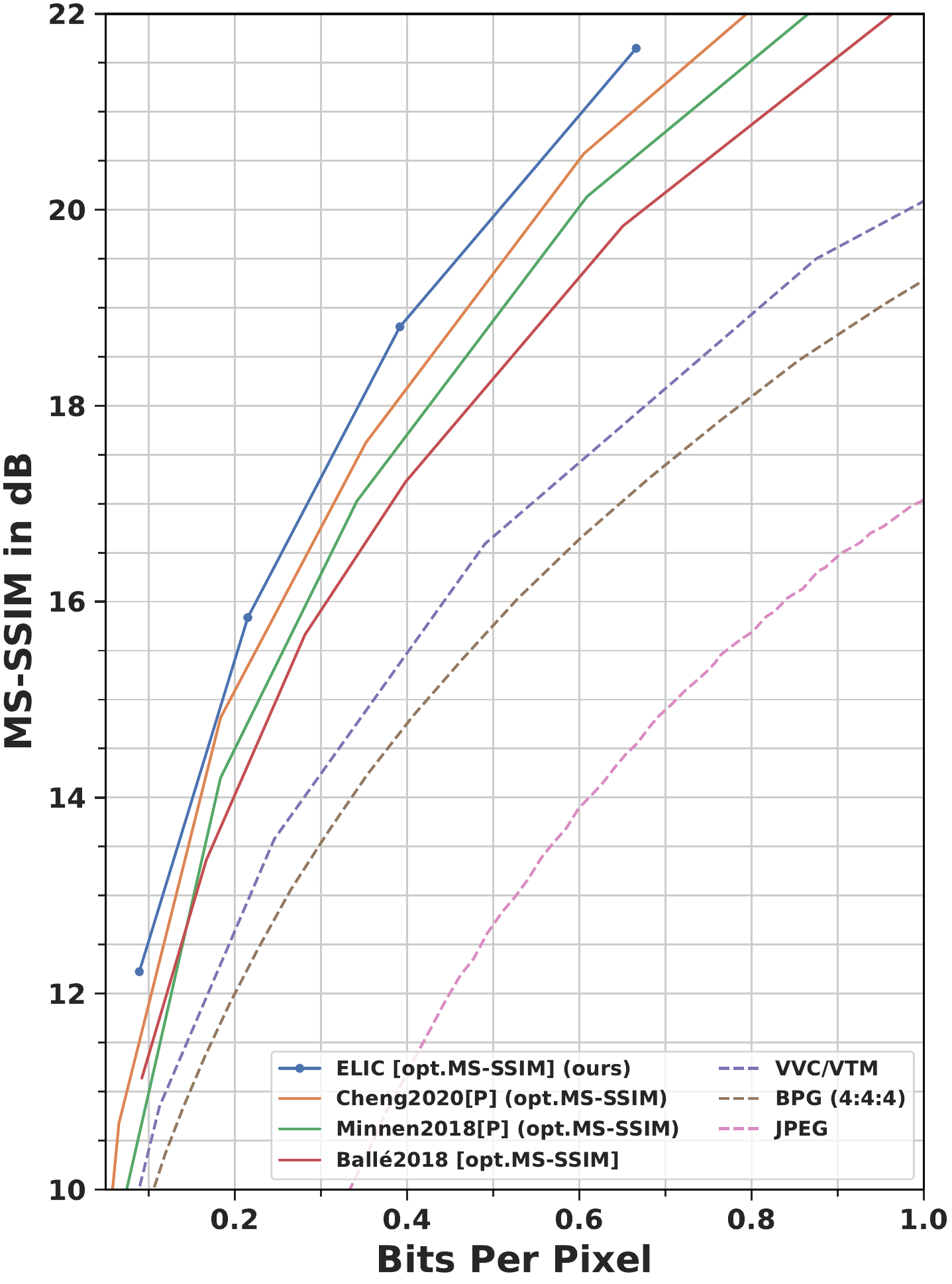}
    \caption{\textbf{MS-SSIM RD curve on Kodak (opt. MS-SSIM).} All models are optimized for maximizing MS-SSIM. All presented learning-based models can decode the $512\times 768$ Kodak image in 100 microseconds.}
    \label{fig:comp-msssim-large-opt-ms}
\end{figure*}

We further evaluate our model performance when optimizing for MS-SSIM, following prior works. Thus, the distortion term of the loss function is replaced by $1 - \mathrm{MSSSIM}(\vx)$. We use $\lambda = \{ 3, 12, 40, 120\}$ to train such models, following Cheng~\etal~(2020). Figure~\ref{fig:comp-msssim-large-opt-ms} shows the results. On Kodak, when achieving the same MS-SSIM, ELIC optimized for MS-SSIM saves about half of the bit-rate compared with VVC, which is encouraging.

\section{Res-blocks v.s. GDN on CPU}

\begin{table}[]
    \centering
    \scalebox{1}{
    \begin{tabular}{rrr}
    \toprule
        \textbf{Layer} & \textbf{YEnc. (s)} &\textbf{YDec. (s)} \\
    \midrule     
         GDN & 0.500 & 1.108  \\
         RB$\times 1$ & 0.448& 1.165 \\
         PWRB$\times 3$ & 0.583 & 1.191 \\
         RB$\times 3$& 0.800 & 1.448 \\
    \bottomrule
    \end{tabular}
    }
    \caption{Ball\'e2018 running speed on CPU with different nonlinear layers, evaluated on Kodak.}
    \label{tab:nonlinear-cpu-comp}
\end{table}

\begin{table}[]
    \centering
    \scalebox{1}{
    \begin{tabular}{r|rr|rr}
    \toprule
        \multirow{3}{*}{\textbf{Layer}} & \multicolumn{2}{c|}{\textbf{YEnc.}} & \multicolumn{2}{c}{\textbf{YDec.}} \\
        & \textbf{Param.} & \textbf{MACs}  & \textbf{Param.} & \textbf{MACs} \\
        & \textbf{(M)} & \textbf{(G)} & \textbf{(M)} & \textbf{(G)} \\
    \midrule     
         GDN &  3.51 & 36.89 & 3.51 & 127.63 \\
         RB$\times 1$ & 3.76 & 47.64 & 3.75 & 138.38  \\
         PWRB$\times 3$& 3.73 & 46.53 & 3.73 & 137.26  \\
         RB$\times 3$ & 4.48 & 78.71 & 4.48 & 169.44 \\
    \bottomrule
    \end{tabular}
    }
    \caption{Ball\'e2018 parameter volumes and MACs with different nonlinear layers, evaluated on Kodak.}
    \label{tab:nonlinear-cpu-param-and-macs}
\end{table}

For completeness, we also evaluate the latency of models with various nonlinear blocks on Intel Core i7-7700 (Table~\ref{tab:nonlinear-cpu-comp}) as well as their parameter volumes and numbers of multiply-accumulate operations (MACs) (Table~\ref{tab:nonlinear-cpu-param-and-macs}). To test parameter volumes and MACs, we use the FLOPs profiler of DeepSpeed~\footnote{\url{https://www.deepspeed.ai/tutorials/flops-profiler/}}. After replacing GDN blocks with residual blocks (RB$\times 1$), the adapted Ball\'e2018 models have slightly larger MACs yet their en/de-coding speeds are on par with the original model with GDNs. This is similar to the GPU results in Table~4 of the main body.

\section{Image reconstruction results}

See Figures~\ref{fig:comp-qualitative-lighthouse},\ref{fig:comp-qualitative-sculpture}, we compare the reconstruction results of the proposed ELIC and Cheng~\etal~(2020), since their synthesis latency is close. We also present the decoded image of VTM12.1 for reference.

\section{Progressive decoding}

In the main text, we present progressive decoding results with thumbnail synthesizer and zero filling. In Figures~\ref{fig:progressive-moto},\ref{fig:progressive-colorhouse}, We further show progressive decoding results with full synthesizer and/or mean filling.

\newcommand{\compquallw}{0.39\linewidth}
\begin{figure*}
    \centering
    \subfloat[Cheng \etal (2020). BPP$=0.129$. PSNR$=29.36$]{
    \includegraphics[width=\compquallw]{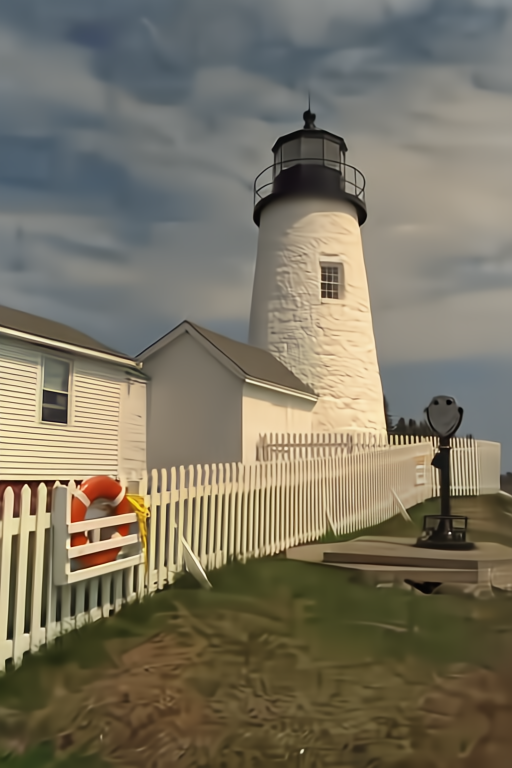}
    }
    \subfloat[Ours. BPP$=0.144$. PSNR$=30.42$]{
    \includegraphics[width=\compquallw]{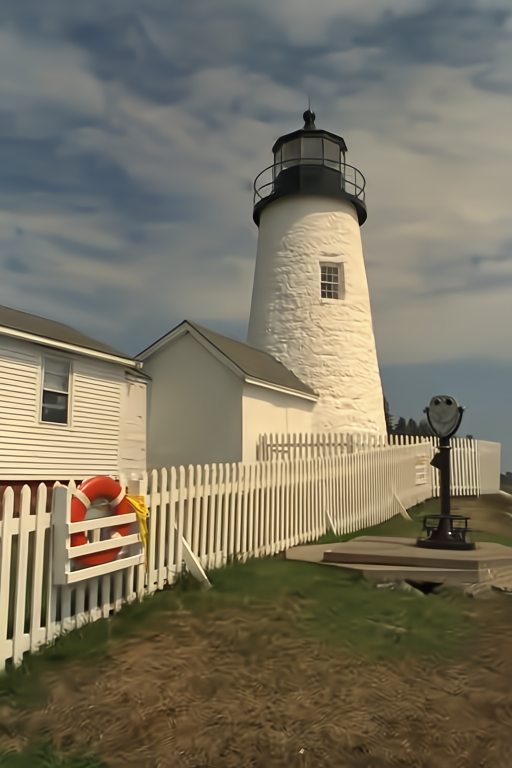}
    }\\
    \subfloat[Ground-truth.]{
    \includegraphics[width=\compquallw]{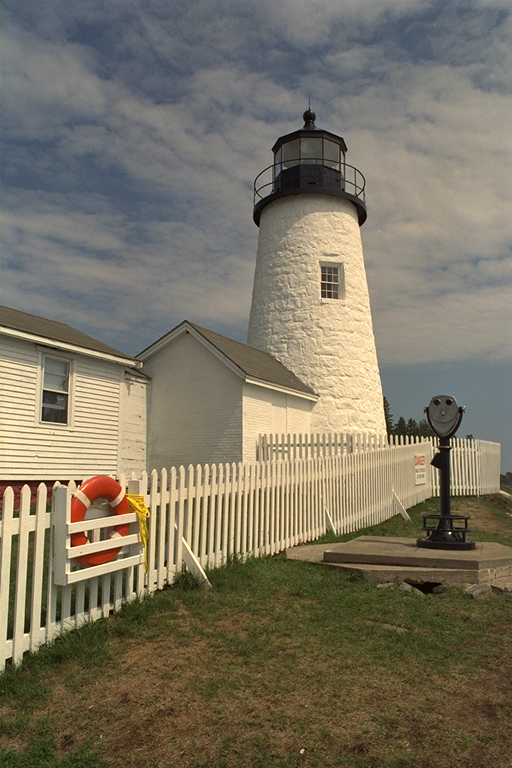}
    }
    \subfloat[VTM12.1. BPP$=0.151$. PSNR$=30.28$]{
    \includegraphics[width=\compquallw]{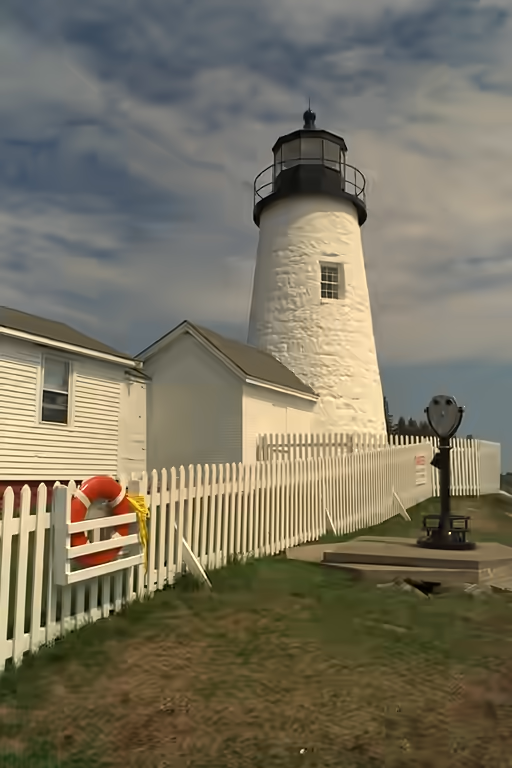}
    }
    \caption{Qualitative comparison on reconstructed lighthouse (\texttt{kodim19}) image.}
    \label{fig:comp-qualitative-lighthouse}
\end{figure*}

\begin{figure*}
    \centering
    \subfloat[Cheng \etal (2020). BPP$=0.155$. PSNR$=29.85$]{
    \includegraphics[width=\compquallw]{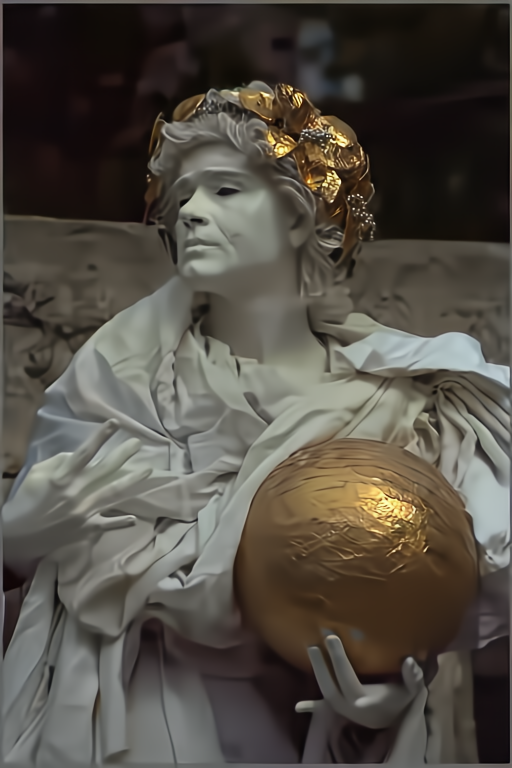}
    }
    \subfloat[Ours. BPP$=0.127$. PSNR$=31.66$]{
    \includegraphics[width=\compquallw]{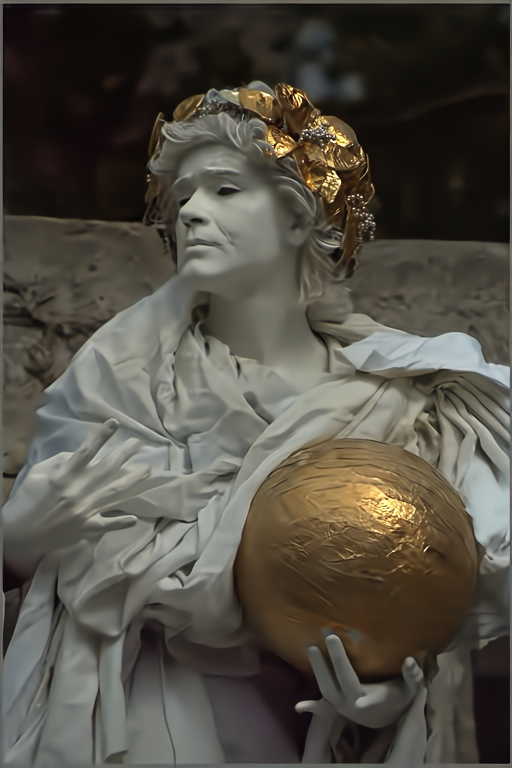}
    }\\
    \subfloat[Ground-truth.]{
    \includegraphics[width=\compquallw]{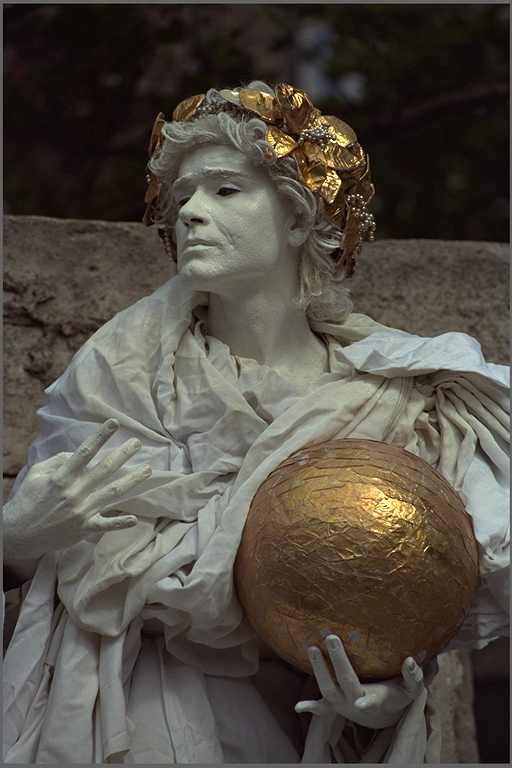}
    }
    \subfloat[VTM12.1. BPP$=0.151$. PSNR$=31.65$]{
    \includegraphics[width=\compquallw]{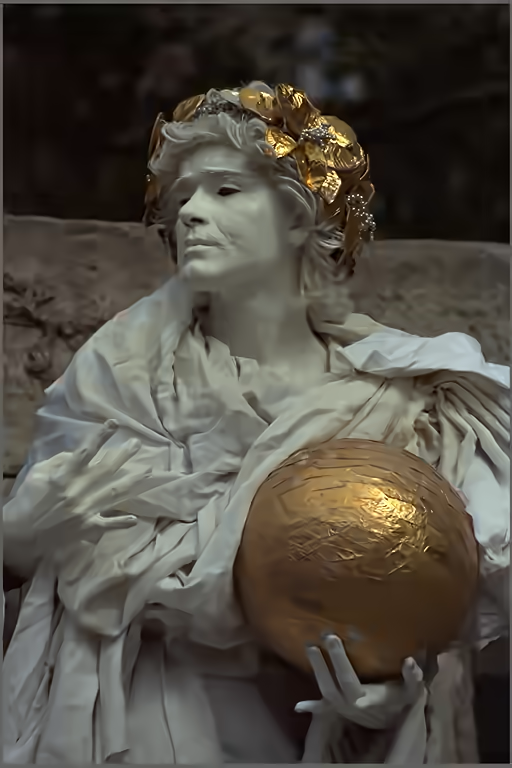}
    }
    \caption{Qualitative comparison on reconstructed sculpture (\texttt{kodim17}) image.}
    \label{fig:comp-qualitative-sculpture}
\end{figure*}

\newcommand{\supprogressivelw}{0.19\linewidth}
\begin{figure*}[htb]
    \centering
    \subfloat{
    \includegraphics[width=\supprogressivelw]{img/progressive/14_reconstruction16.png}
    }
    \subfloat{
    \includegraphics[width=\supprogressivelw]{img/progressive/14_reconstruction32.png}
    }
    \subfloat{
    \includegraphics[width=\supprogressivelw]{img/progressive/14_reconstruction64.png}
    }
    \subfloat{
    \includegraphics[width=\supprogressivelw]{img/progressive/14_reconstruction128.png}
    }
    \subfloat{
    \includegraphics[width=\supprogressivelw]{img/progressive/14_reconstruction320.png}
    }
    \\
    \subfloat{
    \includegraphics[width=\supprogressivelw]{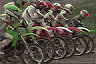}
    }
    \subfloat{
    \includegraphics[width=\supprogressivelw]{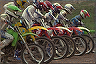}
    }
    \subfloat{
    \includegraphics[width=\supprogressivelw]{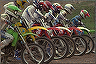}
    }
    \subfloat{
    \includegraphics[width=\supprogressivelw]{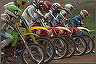}
    }
    \subfloat{
    \includegraphics[width=\supprogressivelw]{img/progressive/14_reconstruction320.png}
    }
    \\
    \subfloat{
    \includegraphics[width=\supprogressivelw]{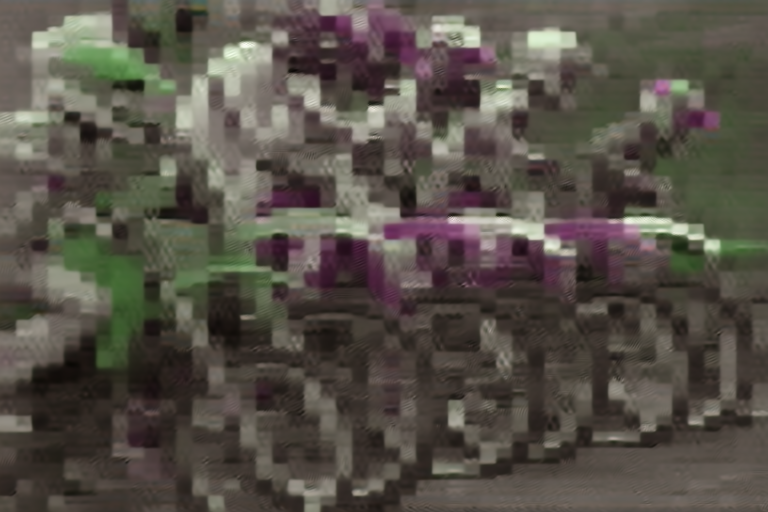}
    }
    \subfloat{
    \includegraphics[width=\supprogressivelw]{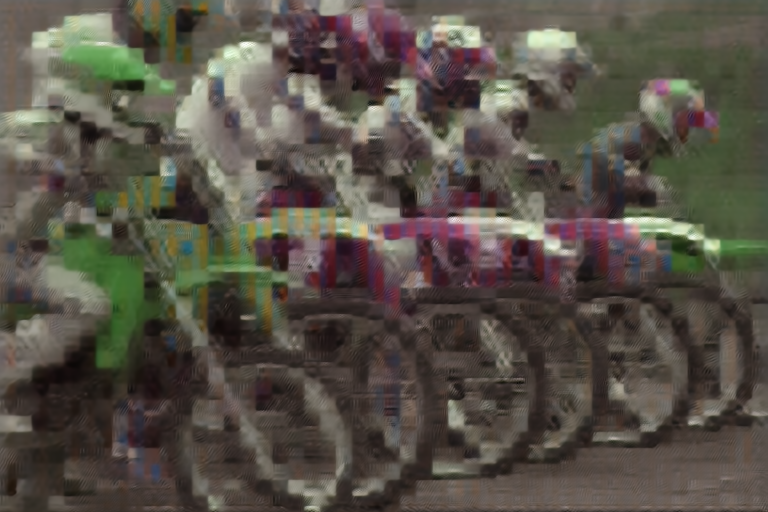}
    }
    \subfloat{
    \includegraphics[width=\supprogressivelw]{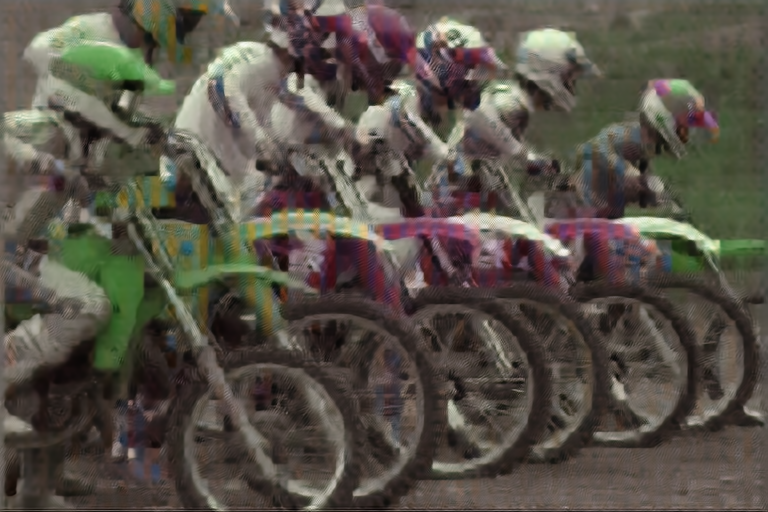}
    }
    \subfloat{
    \includegraphics[width=\supprogressivelw]{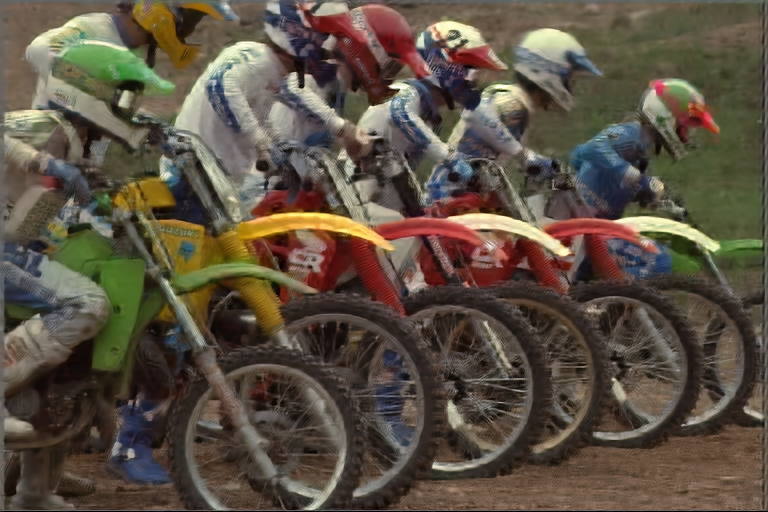}
    }
    \subfloat{
    \includegraphics[width=\supprogressivelw]{img/progressive/14_reconstruction320.png}
    }
    \\
    \subfloat{
    \includegraphics[width=\supprogressivelw]{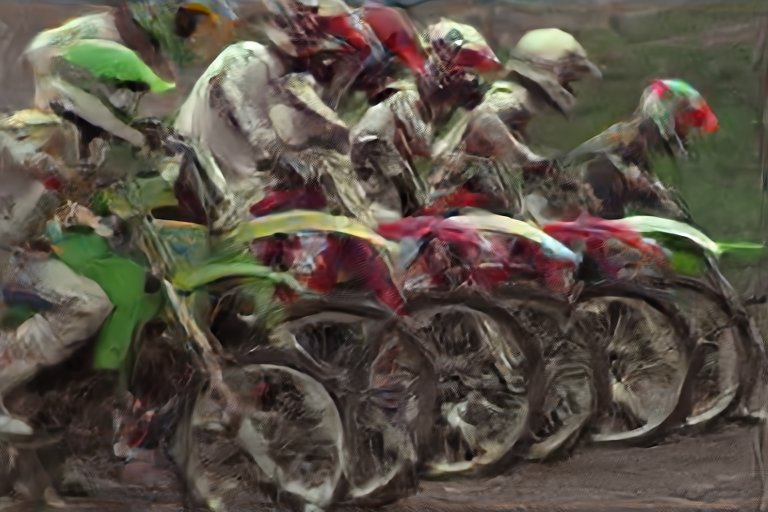}
    }
    \subfloat{
    \includegraphics[width=\supprogressivelw]{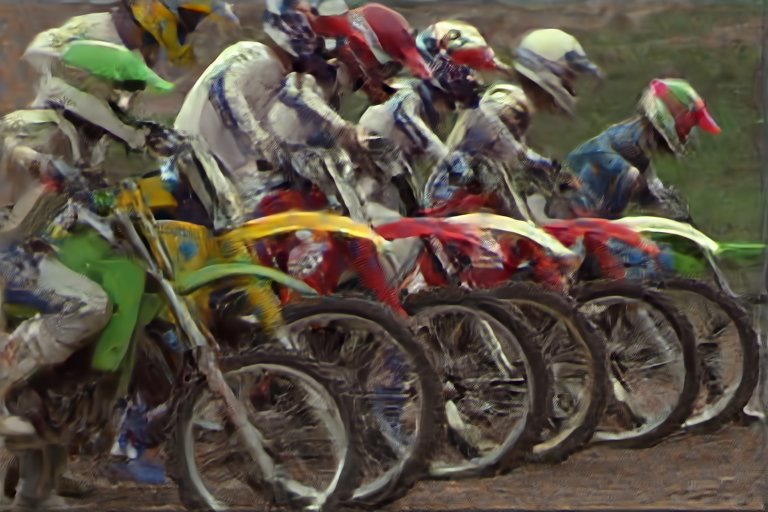}
    }
    \subfloat{
    \includegraphics[width=\supprogressivelw]{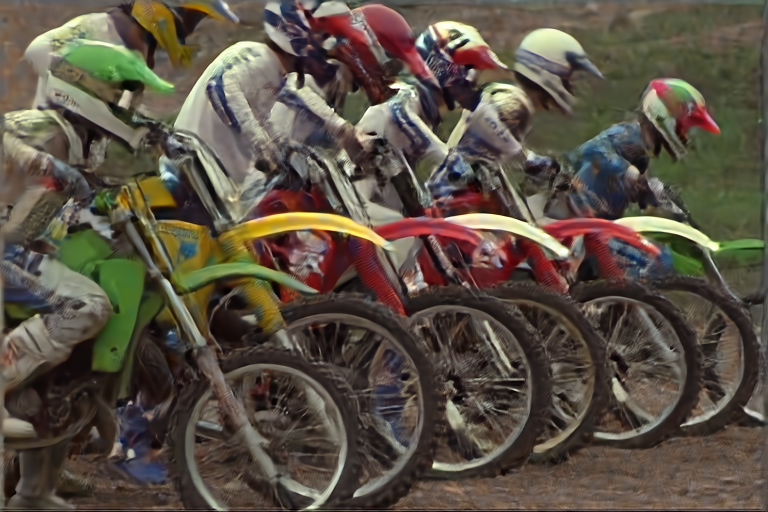}
    }
    \subfloat{
    \includegraphics[width=\supprogressivelw]{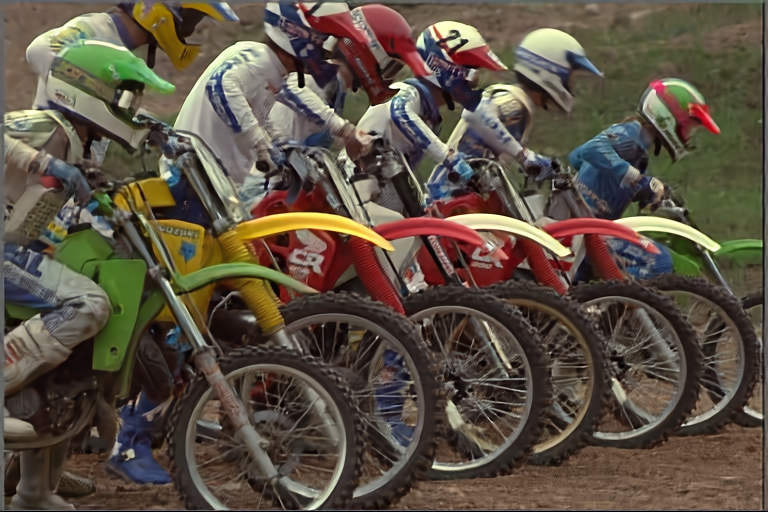}
    }
    \subfloat{
    \includegraphics[width=\supprogressivelw]{img/progressive/14_reconstruction320.png}
    }
    \caption{Progressive decoding results (\texttt{kodim05}). \\Line 1 (thumb-zero): the first 4 groups are decoded with thumbnail decoder and zero filling (the same setting as the main text). \\Line 2 (thumb-mean): the thumbnail-decoding results with mean filling. \\Line 3 (full-zero): all the 5 groups are decoded with full synthesizer, using zero filling. \\Line 4 (full-mean): the full-decoding results with mean filling.}
    \label{fig:progressive-moto}
\end{figure*}

\begin{figure*}[htb]
    \centering
    \subfloat{
    \includegraphics[width=\supprogressivelw]{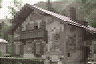}
    }
    \subfloat{
    \includegraphics[width=\supprogressivelw]{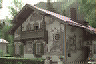}
    }
    \subfloat{
    \includegraphics[width=\supprogressivelw]{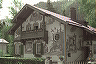}
    }
    \subfloat{
    \includegraphics[width=\supprogressivelw]{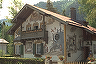}
    }
    \subfloat{
    \includegraphics[width=\supprogressivelw]{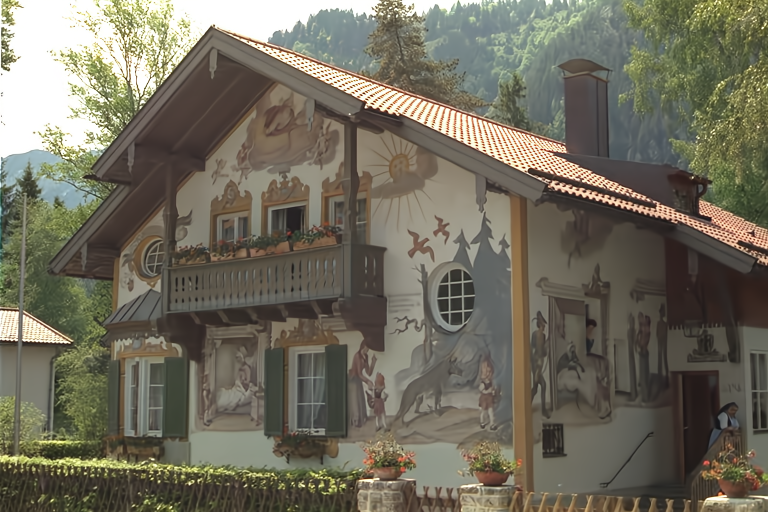}
    }
    \\
    \subfloat{
    \includegraphics[width=\supprogressivelw]{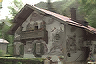}
    }
    \subfloat{
    \includegraphics[width=\supprogressivelw]{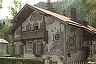}
    }
    \subfloat{
    \includegraphics[width=\supprogressivelw]{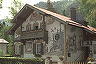}
    }
    \subfloat{
    \includegraphics[width=\supprogressivelw]{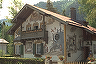}
    }
    \subfloat{
    \includegraphics[width=\supprogressivelw]{img/progressive/11_reconstruction320.png}
    }
    \\
    \subfloat{
    \includegraphics[width=\supprogressivelw]{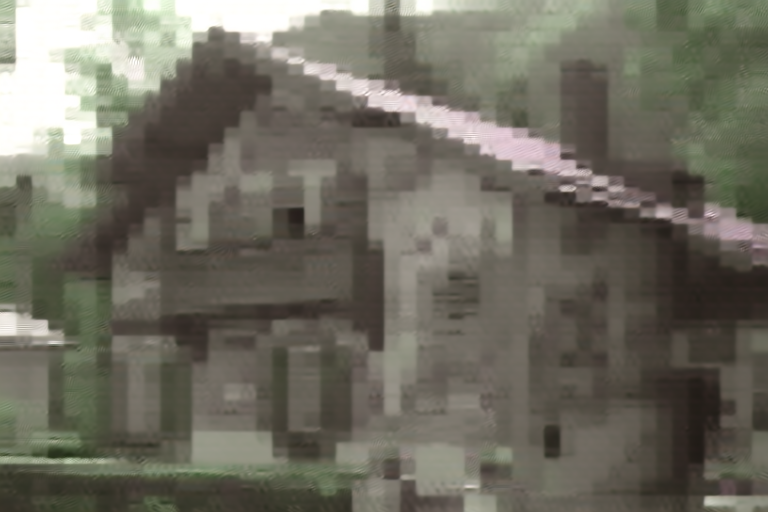}
    }
    \subfloat{
    \includegraphics[width=\supprogressivelw]{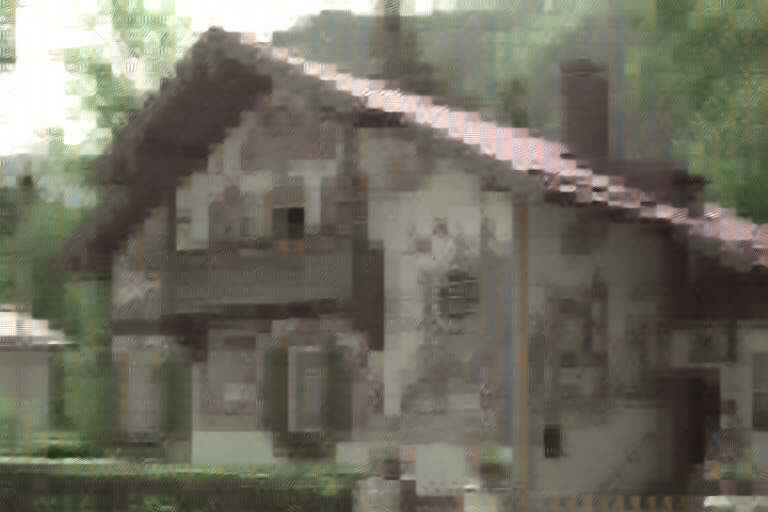}
    }
    \subfloat{
    \includegraphics[width=\supprogressivelw]{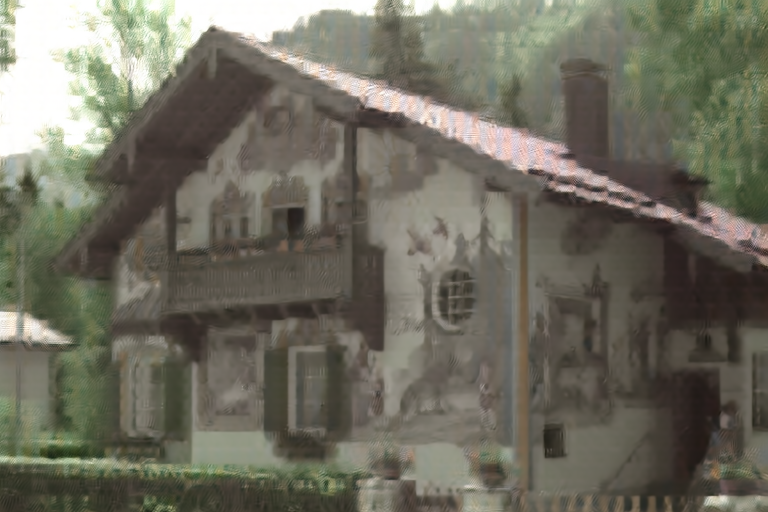}
    }
    \subfloat{
    \includegraphics[width=\supprogressivelw]{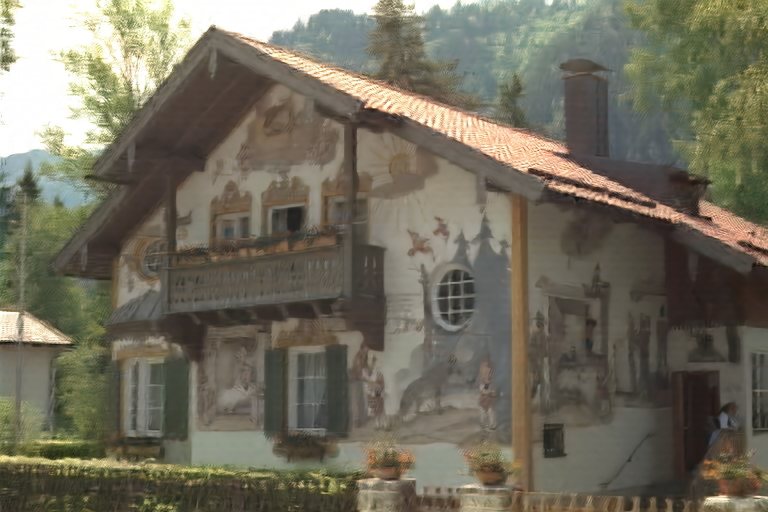}
    }
    \subfloat{
    \includegraphics[width=\supprogressivelw]{img/progressive/11_reconstruction320.png}
    }
    \\
    \subfloat{
    \includegraphics[width=\supprogressivelw]{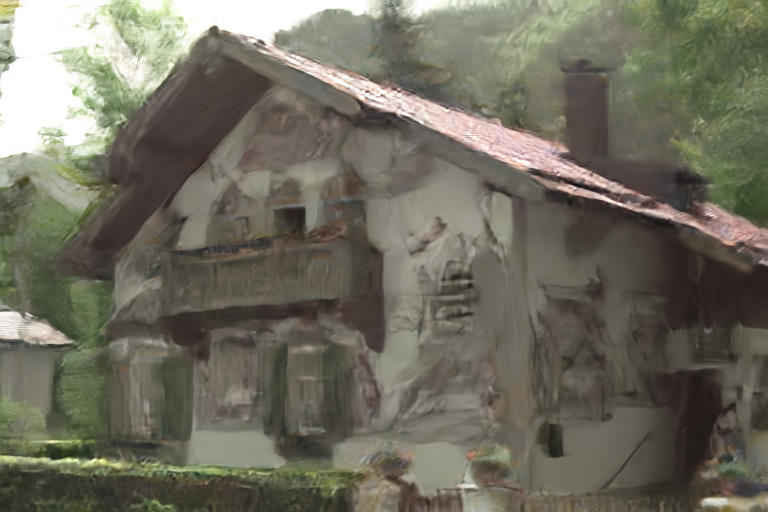}
    }
    \subfloat{
    \includegraphics[width=\supprogressivelw]{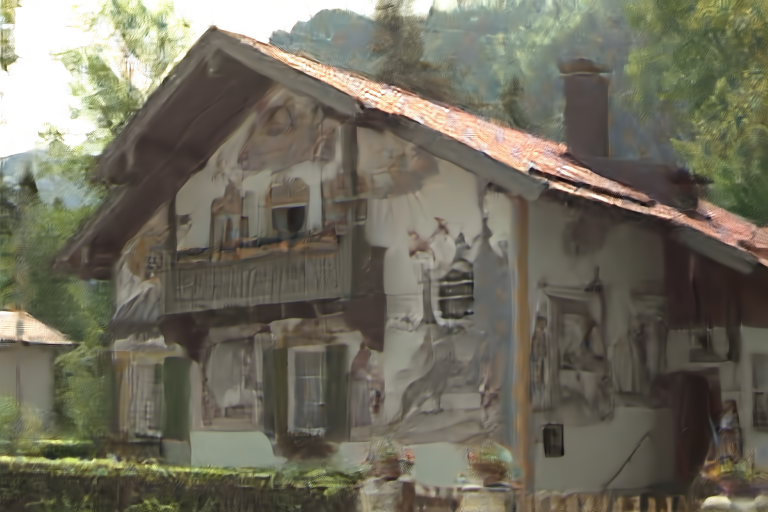}
    }
    \subfloat{
    \includegraphics[width=\supprogressivelw]{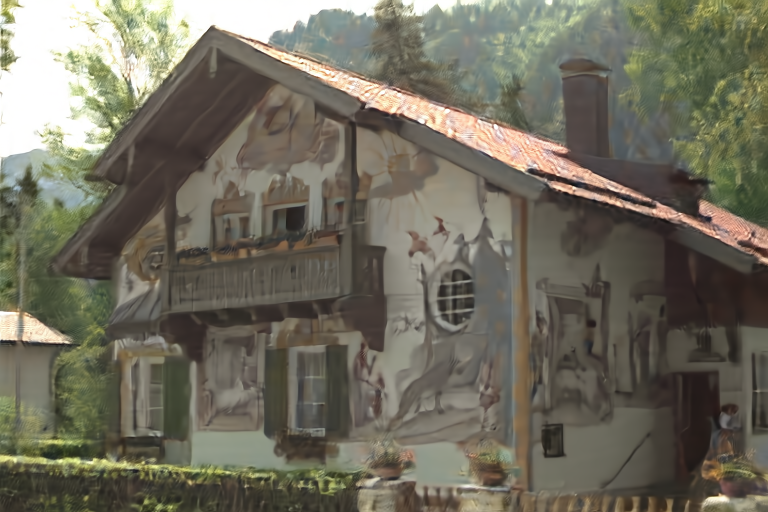}
    }
    \subfloat{
    \includegraphics[width=\supprogressivelw]{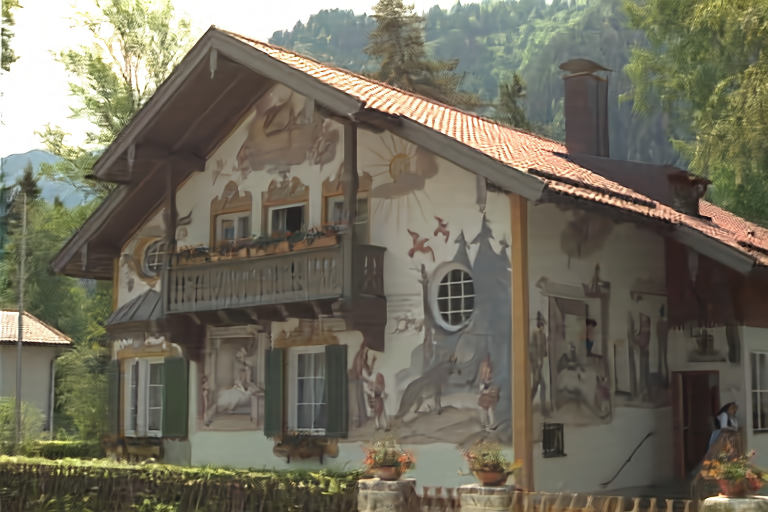}
    }
    \subfloat{
    \includegraphics[width=\supprogressivelw]{img/progressive/11_reconstruction320.png}
    }
    \caption{Progressive decoding results of the house image (\texttt{kodim24}). The image order is the same as Figure~\ref{fig:progressive-moto}.}
    \label{fig:progressive-colorhouse}
\end{figure*}